\documentclass[10pt,journal,compsoc]{IEEEtran}
\usepackage[pagebackref=true,breaklinks=true,citecolor=blue,linkcolor=blue,urlcolor=blue,colorlinks,bookmarks=false]{hyperref}
\usepackage{booktabs}
\renewcommand{\arraystretch}{1}

\ifCLASSOPTIONcompsoc
  \usepackage[nocompress]{cite}
\else
  \usepackage{cite}
\fi
\ifCLASSINFOpdf
\else
\fi
\usepackage{graphicx} 
\usepackage{amsmath}

\def\etal{\textit{et al.}}
\usepackage{pgf-pie}
\usepackage{pgfplots}
\usetikzlibrary{patterns}
\usetikzlibrary{arrows,shapes,positioning,shadows,trees}
\usepackage{multirow, tabularx}
\usepackage{makecell}

\usepackage{xcolor,colortbl}

\newcolumntype{a}{>{\columncolor{white}}c}
\pgfplotsset{compat=1.3}

\tikzset{
  basic/.style  = {draw, text width=4cm, drop shadow, font=\sffamily, rectangle},
  root/.style   = {basic, rectangle, very thick, align=center,
                  fill=blue!60},
  level 2/.style = {basic, rectangle, very thick, align=center, fill=gray!30,
                    text width=12em},
  level 3/.style = {basic, very thick, align=left, fill=orange!40, text width=9em}
}

\newcommand*{\belowrulesepcolor}[1]{% 
  \noalign{% 
    \kern-\belowrulesep 
    \begingroup 
      \color{#1}% 
      \hrule height\belowrulesep 
    \endgroup 
    \vspace{-0.03mm}
  }%
} 
\newcommand*{\aboverulesepcolor}[1]{% 
  \noalign{% 
  \vspace{-0.03mm}
    \begingroup 
      \color{#1}% 
      \hrule height\aboverulesep 
    \endgroup 
    \kern-\aboverulesep 
  }%
}

\newcommand{\vspacefigtext}{\vspace{-3mm}}
\newcommand{\vspacesection}{\vspace{-2.2mm}}
\renewcommand{\IEEEbibitemsep}{0pt plus 0.5pt}
\makeatletter
\IEEEtriggercmd{\reset@font\normalfont\fontsize{7.3pt}{8.40pt}\selectfont}
\makeatother
\IEEEtriggeratref{1}

\begin{document}
%
% paper title
% Titles are generally capitalized except for words such as a, an, and, as,
% at, but, by, for, in, nor, of, on, or, the, to and up, which are usually
% not capitalized unless they are the first or last word of the title.
% Linebreaks \\ can be used within to get better formatting as desired.
% Do not put math or special symbols in the title.
\title{3D Vision with Transformers: A Survey}
%
%
% author names and IEEE memberships
% note positions of commas and nonbreaking spaces ( ~ ) LaTeX will not break
% a structure at a ~ so this keeps an author's name from being broken across
% two lines.
% use \thanks{} to gain access to the first footnote area
% a separate \thanks must be used for each paragraph as LaTeX2e's \thanks
% was not built to handle multiple paragraphs
%
%
%\IEEEcompsocitemizethanks is a special \thanks that produces the bulleted
% lists the Computer Society journals use for "first footnote" author
% affiliations. Use \IEEEcompsocthanksitem which works much like \item
% for each affiliation group. When not in compsoc mode,
% \IEEEcompsocitemizethanks becomes like \thanks and
% \IEEEcompsocthanksitem becomes a line break with idention. This
% facilitates dual compilation, although admittedly the differences in the
% desired content of \author between the different types of papers makes a
% one-size-fits-all approach a daunting prospect. For instance, compsoc 
% journal papers have the author affiliations above the "Manuscript
% received ..."  text while in non-compsoc journals this is reversed. Sigh.

\author{Jean~Lahoud,
        Jiale~Cao,
        Fahad~Shahbaz~Khan,
        Hisham~Cholakkal,
        Rao~Muhammad~Anwer,
        Salman~Khan,    
        and~Ming-Hsuan~Yang% <-this % stops a space
\IEEEcompsocitemizethanks{\IEEEcompsocthanksitem J. Lahoud, F. Khan, H. Cholakkal, R. Anwer, and S. Khan are with Mohamed bin Zayed University of Artificial Intelligence, UAE.\protect\\
E-mail: \{firstname.lastname\}@mbzuai.ac.ae
% \IEEEcompsocthanksitem J. Cao is with the School of Electrical and Information Engineering, Tianjin University, Tianjin 300072, China \protect\\
\IEEEcompsocthanksitem J. Cao is with the School of Electrical and Information Engineering, Tianjin University, China. 
E-mail: connor@tju.edu.cn
\IEEEcompsocthanksitem F. Khan is also with Link\"{o}ping University, Sweden
\IEEEcompsocthanksitem S. Khan is also with Australian National University, Australia
% \IEEEcompsocthanksitem M.-H. Yang is with the Department of Computer Science and Engineering, University of California, Merced, CA, 95340, USA.\protect\\
\IEEEcompsocthanksitem M.-H. Yang is with University of California at Merced, Yonsei University, and Google.
Email:mhyang@ucmerced.edu}% <-this % stops an unwanted space

% \IEEEcompsocthanksitem J. Doe and J. Doe are with Anonymous University.}% <-this % stops an unwanted space
% \IEEEcompsocitemizethanks{\IEEEcompsocthanksitem M. Shell was with the Department
% of Electrical and Computer Engineering, Georgia Institute of Technology, Atlanta,
% GA, 30332.\protect\\
% % note need leading \protect in front of \\ to get a newline within \thanks as
% % \\ is fragile and will error, could use \hfil\break instead.
% E-mail: see http://www.michaelshell.org/contact.html
% \IEEEcompsocthanksitem J. Doe and J. Doe are with Anonymous University.}% <-this % stops an unwanted space
% \thanks{Manuscript received April 19, 2005; revised August 26, 2015.}
}

% note the % following the last \IEEEmembership and also \thanks - 
% these prevent an unwanted space from occurring between the last author name
% and the end of the author line. i.e., if you had this:
% 
% \author{....lastname \thanks{...} \thanks{...} }
%                     ^------------^------------^----Do not want these spaces!
%
% a space would be appended to the last name and could cause every name on that
% line to be shifted left slightly. This is one of those "LaTeX things". For
% instance, "\textbf{A} \textbf{B}" will typeset as "A B" not "AB". To get
% "AB" then you have to do: "\textbf{A}\textbf{B}"
% \thanks is no different in this regard, so shield the last } of each \thanks
% that ends a line with a % and do not let a space in before the next \thanks.
% Spaces after \IEEEmembership other than the last one are OK (and needed) as
% you are supposed to have spaces between the names. For what it is worth,
% this is a minor point as most people would not even notice if the said evil
% space somehow managed to creep in.

% The paper headers
% \markboth{IEEE Transactions on Pattern Analysis and Machine Intelligence}%
{}
% The only time the second header will appear is for the odd numbered pages
% after the title page when using the twoside option.
% 
% *** Note that you probably will NOT want to include the author's ***
% *** name in the headers of peer review papers.                   ***
% You can use \ifCLASSOPTIONpeerreview for conditional compilation here if
% you desire.

% The publisher's ID mark at the bottom of the page is less important with
% Computer Society journal papers as those publications place the marks
% outside of the main text columns and, therefore, unlike regular IEEE
% journals, the available text space is not reduced by their presence.
% If you want to put a publisher's ID mark on the page you can do it like
% this:
%\IEEEpubid{0000--0000/00\$00.00~\copyright~2015 IEEE}
% or like this to get the Computer Society new two part style.
%\IEEEpubid{\makebox[\columnwidth]{\hfill 0000--0000/00/\$00.00~\copyright~2015 IEEE}%
%\hspace{\columnsep}\makebox[\columnwidth]{Published by the IEEE Computer Society\hfill}}
% Remember, if you use this you must call \IEEEpubidadjcol in the second
% column for its text to clear the IEEEpubid mark (Computer Society jorunal
% papers don't need this extra clearance.)

% use for special paper notices
%\IEEEspecialpapernotice{(Invited Paper)}

% for Computer Society papers, we must declare the abstract and index terms
% PRIOR to the title within the \IEEEtitleabstractindextext IEEEtran
% command as these need to go into the title area created by \maketitle.
% As a general rule, do not put math, special symbols or citations
% in the abstract or keywords.
\IEEEtitleabstractindextext{%
\begin{abstract}
The success of the transformer architecture in natural language processing has recently triggered attention in the computer vision field. 
The transformer has been used as a replacement for the widely used convolution operators, due to its ability to learn long-range dependencies. 
This replacement was proven to be successful in numerous tasks, in which several state-of-the-art methods rely on transformers for better learning. 
In computer vision, the 3D field has also witnessed an increase in employing the transformer for 3D convolution neural networks and multi-layer perceptron networks. 
Although a number of surveys have focused on transformers in vision in general, 3D vision requires special attention due to the difference in data representation and processing when compared to 2D vision. 
In this work, we present a systematic and thorough review of more than 100 transformers methods for different 3D vision tasks, including classification, segmentation, detection, completion, pose estimation, and others.
%systematic survey on transformers for 3D vision tasks, including classification, segmentation, detection, completion, pose estimation, and others.
We discuss transformer design in 3D vision, which allows it to process data with various 3D representations. 
For each application, we highlight key properties and contributions of proposed transformer-based methods. 
To assess the competitiveness of these methods, we compare their performance to common non-transformer methods on 12 3D benchmarks. 
We conclude the survey by discussing different open directions and challenges for transformers in 3D vision.
In addition to the presented papers, we aim to frequently update the latest relevant papers along with their corresponding implementations at: \href{https://github.com/lahoud/3d-vision-transformers}{https://github.com/lahoud/3d-vision-transformers}.
\end{abstract}

% Note that keywords are not normally used for peerreview papers.
}

% make the title area
\maketitle

% To allow for easy dual compilation without having to reenter the
% abstract/keywords data, the \IEEEtitleabstractindextext text will
% not be used in maketitle, but will appear (i.e., to be "transported")
% here as \IEEEdisplaynontitleabstractindextext when the compsoc 
% or transmag modes are not selected <OR> if conference mode is selected 
% - because all conference papers position the abstract like regular
% papers do.
\IEEEdisplaynontitleabstractindextext
% \IEEEdisplaynontitleabstractindextext has no effect when using
% compsoc or transmag under a non-conference mode.

% For peer review papers, you can put extra information on the cover
% page as needed:
% \ifCLASSOPTIONpeerreview
% \begin{center} \bfseries EDICS Category: 3-BBND \end{center}
% \fi
%
% For peerreview papers, this IEEEtran command inserts a page break and
% creates the second title. It will be ignored for other modes.
\IEEEpeerreviewmaketitle

\IEEEraisesectionheading{\section{Introduction}\label{sec:introduction}}
% \IEEEPARstart{T}{his} demo file is intended to serve as a ``starter file''
% for IEEE Computer Society journal papers produced under \LaTeX\ using
% IEEEtran.cls version 1.8b and later.
% You must have at least 2 lines in the paragraph with the drop letter
% (should never be an issue)

% Importance of 3d vision
%\IEEEraisesectionheading{\section{Introduction}}
%\IEEEPARstart{3}{D}
%vision aims to represent and understand the world in three dimensions, which is the number of dimensions in the real physical space. 
\IEEEPARstart{O}{ne} fundamental problem in computer vision is to understand the scenes and objects in the three-dimension space. 
It allows a compact representation of relationships and provides the ability to navigate and manipulate in the real world. 3D vision plays an important role in various domains and includes applications in autonomous driving, robotics, remote sensing, medical treatment, augmented reality, design industry, among many others. 
There has been an increasing interest in the 3D field due to numerous reasons: (1) development of various 3D capture sensors, e.g., LiDAR and RGB-D sensors, (2) introduction of numerous large-scale 3D geometry datasets that were collected and labeled in 3D, and (3) advances in 3D deep learning methods. 

Common approaches to 3D deep learning methods employ deep convolution neural networks (CNNs) and multilayer perceptrons (MLPs).
Nevertheless, transformer-based architectures that use the attention mechanism have shown a strong contender for such methods in various fields, such as natural language processing (NLP) and 2D image processing. 
While convolution operators have limited receptive fields and translation equivariance properties, the attention mechanism operates globally and can thus encode long-range dependencies, allowing attention-based methods to learn richer feature representations. 
% can add examples of successful transformer methods

Witnessing the success of transformer-based architectures in the image domain, numerous 3D vision methods have recently adopted  transformers in the model designs. 
These architectures have been proposed as a solution for most common 3D vision applications. 
In 3D, the transformer has replaced previous learning methods or supplemented them, benefiting from its ability to capture long-range information and learn task-specific inductive biases.

\pgfplotstableread[row sep=\\,col sep=&]{
    interval & num  \\
    Jun18-Mar19     & 1   \\
    Apr19-Jan20     & 3  \\
    Feb20-Oct20    & 10  \\
    Nov20-Aug21   & 32 \\
    Sep21-Jun22   & 61   \\
    }\mydata

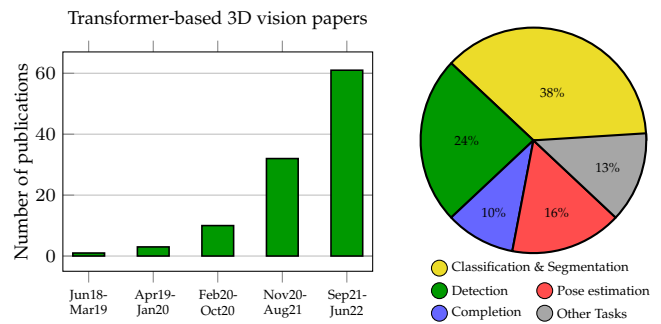
\begin{figure}[t]
\renewcommand{\arraystretch}{1.3}
\centering
\begin{tabular}{a a}
\begin{tikzpicture}
\scriptsize
\begin{axis} [ybar,height=4.5cm,width=5.7cm,
symbolic x coords={Jun18-Mar19,Apr19-Jan20,Feb20-Oct20,Nov20-Aug21,Sep21-Jun22},
x tick label style={font=\tiny,text width=0.5cm,align=center},
xtick=data,
% xlabel=Date of publication,
title = Transformer-based 3D vision papers,
ylabel shift = -4pt,
ylabel=Number of publications,
ymajorgrids=true,
yminorgrids=true]
\addplot [draw = black,
    fill=green!60!black,
    semithick,
    % pattern = dots,
    % pattern color = black,
    bar width = 12
] table[x=interval,y=num]{\mydata};;
\end{axis}
\end{tikzpicture} &
\begin{tikzpicture}[align=center]]
\tiny
\pie[radius=1.5,
    color = {
        yellow!90!black,
        green!60!black,
        blue!60, red!70,
        gray!70,
        teal!20},%,
        %red!20}
]{  38/~,
    24/~,
    10/~, 
    16/~,
    13/~};
    % 1.07/
% }
\draw (-1.25,-1.7) [fill=yellow!90!black] circle (3pt) node [right=0.05]{~Classification \& Segmentation}; 
\draw (-1.25,-2) [fill=green!60!black] circle (3pt) node [right=0.05]{~Detection};
\draw (-1.25,-2.3) [fill=blue!60] circle (3pt) node [right=0.05]{~Completion};
\draw (0.15,-2) [fill=red!70] circle (3pt) node [right=0.05]{~Pose estimation};
\draw (0.15,-2.3) [fill=gray!70] circle (3pt) node [right=0.05]{~Other Tasks};

\end{tikzpicture}
\end{tabular}
\caption{(Left) A bar graph showing the number of transformer-based 3D vision papers in recent years. For papers available on arXiv, we use the first submission date; otherwise, the journal or conference publication date is used. A consistent growth reflects the increased attention to the transformer architecture in the recent literature. (Right) A pie chart showing the percentage of papers in this survey for each application.}
\label{fig:stat}
\vspacefigtext
\end{figure}

Given the increasing interest of transformers in 3D vision (Fig. \ref{fig:stat}, left), a survey that gives an overview of the available methods is of great importance to give a holistic view of this emerging field. 
In this survey, we review methods that use transformers for 3D vision tasks, including classification, segmentation, detection, completion, pose estimation, and others (Fig. \ref{fig:stat}, right). 
We highlight transformer design choices in 3D vision, which allows
it to process data with various 3D representations. 
For each application, we discuss key properties and contributions of proposed
transformer-based methods. 
Finally, we compare their performance to alternative methods on the widely used 3D datasets/benchmarks to assess the competitiveness of the transformer integration in this field.

We note that numerous surveys have studied deep learning methods in 3D vision. 
Among these surveys, many studies that have been published provide an overall review of methods that process 3D data \cite{guo2020deep,ioannidou2017deep,bello2020deep,liu2019deep}.
Other studies focus on specific 3D vision applications, such as segmentation \cite{he2021deep,gao2021we,xie2020linking}, classification \cite{griffiths2019review}, or detection \cite{fernandes2021point,wu2020deep}.
Furthermore, some surveys examine 3D deep learning methods from a representation perspective \cite{xiao2020survey,ahmed2018survey}, and others limit their studies to a specific data input sensor \cite{wu2020deep,li2020deep}.
Given that most of the surveys were published prior to the recent success of the transformer architecture, attention to the transformer-based architectures is still missing.

% surveys for transformers
With the plethora of recent vision methods that rely on the attention mechanism and the transformer architecture, many works have emerged that survey these methods. Some of these works consider transformers in vision in general \cite{khan2021transformers,liu2021survey,han2022survey,lin2021survey,guo2022attention}, while others focus on a specific aspect, such as efficiency \cite{tay2020efficient}, or a specific application, such as video \cite{shamshad2022transformers} or medical imaging \cite{selva2022video}.
Considering the differences between the 2D and 3D data representation and processing, special attention to transformers applied to 3D vision applications is essential. 
Thus, we focus on transformer architectures applied in the 3D vision field.

%\subsection{Scope}
This survey includes methods that employ the transformer architecture with 3D inputs and/or outputs. 
3D data can be obtained with numerous sensors, such as RGB-D sensors for indoors, LiDAR for outdoors, as well as specialized medical sensors. We include methods that either use point clouds as input or dense 3D grid.
A dense 3D grid can also be obtained by taking images at different slices, which is common in medical imaging. 
In addition, representative methods that apply the transformer architectures to other input data, such as multi-view images or bird-eye view images, and generate output in 3D are also included.

\vspacesection
\section{Preliminaries}\label{sec:background}
Significant advances have been recently made in the field of 3D computer vision. 
%Methods use a variety of 3D representations and processing techniques. 
%
In this section, we first review different representations of 3D data, as well as numerous processing techniques that enable learning from such data. 
For the transformer model, we present its main component (attention), architecture, and valuable properties.

\vspacesection
\subsection{3D Representation}
\begin{figure}[ht]
\renewcommand{\arraystretch}{1.3}
\centering
\footnotesize
\begin{tabular}{c}
\vspace{-2mm}\includegraphics[width=0.7\linewidth]{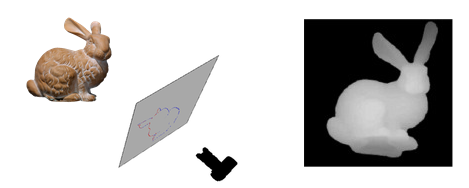} \\
Projection-based 3D representation (Depth Image)\\
\includegraphics[width=0.7\linewidth]{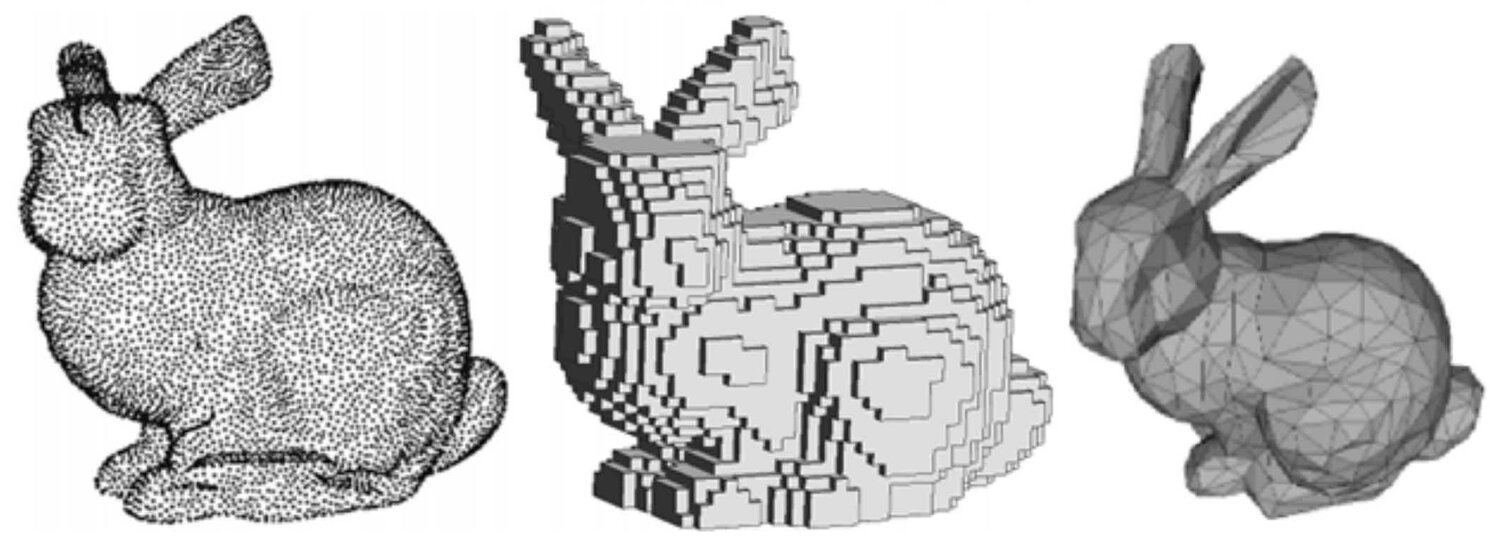} \\
Point Cloud \hspace{1cm} Voxels \hspace{12mm} Mesh  \\
\end{tabular}
\caption{Various 3D representations of the Stanford bunny \cite{turk1994zippered}. The projection-based 3D representation can only visualize an object from one viewpoint. Other representations provide 3D information for all viewpoints.}
\label{fig:3D_rep}
\vspacefigtext
\end{figure}

Images and videos have an inherent natural representation characterized by pixels on a standard grid. 
On the other hand, such organized grid structure does not exist for 3D geometry. 
In this section, we discuss widely-used representations of 3D data which allow employing different deep learning algorithms/techniques. Fig. \ref{fig:3D_rep} shows different 3D representations of the Stanford bunny.% for the goal of understanding such data. 

% \begin{figure}[h]
% \centering
% \includegraphics[width=\linewidth]{images/rgbd.png}
% \caption{Depth image of Stanford bunny}
% \label{fig:3d_rep}
% \end{figure}

%\subsubsection{Multi-view Representation}
\vspace{1mm}
\noindent \textbf{Multi-view Representation.}
A 3D shape can be represented by a set of 2D images captured from different viewpoints. 
Compared to other representations in 3D, this representation is relatively efficient mainly due to having one less dimension, which yields a smaller data size. 
With this representation, one can exploit 2D learning methods for 3D analysis.
Capturing such data is readily obtainable using 2D cameras, as opposed to 3D sensors that are more expensive and less common.
Although multi-view representation targets easier 2D processing, one can also extract 3D information and process it in 3D. 
This is carried out through stereo vision in which the relative positions of objects in multiple views allow the extraction of 3D information using the triangulation of the camera rays. 

%\subsubsection{Depth Images}
\vspace{1mm}
\noindent \textbf{Depth Images.}
A depth image provides the distance between the camera and the scene for each pixel (Fig. \ref{fig:3D_rep}). 
Such data is commonly presented as RGB-D data, which is a structured representation that is constituted of a color image and a corresponding depth image. 
%
%MH: Is this correct? I comment this one out. My understanding of %2.5D is not this. Check the definition in textbooks at   %https://en.wikipedia.org/wiki/2.5D_(visual_perception)  %https://en.wikipedia.org/wiki/2.5D
%jean: I think the terminology is used since depth images share similarities with the 2.5D visual perception
% https://www.cv-foundation.org/openaccess/content_cvpr_2015/papers/Wu_3D_ShapeNets_A_2015_CVPR_paper.pdf
% I have kept the following commented as it might cause confusion.
%As RGB-D data is captured from one viewpoint, it is referred to as 2.5D. 
%
This data can be easily acquired using depth sensors such as the Kinect, among many others. % Microsoft Kinect
Depth images can also be obtained from multi-view/stereo images, in which a disparity map is calculated for every pixel within an image. 
Since a depth image is captured from one viewpoint, it does not describe the whole object geometry - objects are seen from one side only. 
Nevertheless, since many 2D algorithms can be directly employed on such structured data, using this representation benefits from the great advances in 2D processing. % since similar algorithms can be applied. %Also, the depth information maps to a pixel location, which help in relating the color information to the 3D location.

%\subsubsection{Point Cloud}
\vspace{1mm}
\noindent \textbf{Point Cloud.}
A point cloud is a set of vertices in 3D space, represented by their coordinates along the x, y, and z axes. 
Such data can be acquired from 3D scanners, e.g., LiDARs or RGB-D sensors, from one or more viewpoints. 
Color information captured by RGB cameras can be optionally superimposed on the point cloud as additional information. 
Unlike images that are usually represented as matrices, a point cloud is an unordered set. 
As such, processing such data entails a permutation invariant method so that the output does not vary with different ordering of the same point cloud.

% \begin{figure}[h]
% \centering
% \includegraphics[width=\linewidth]{images/3D_representations.jpg}
% \caption{Different representations of the 3D Stanford bunny. (left) Point cloud representation, (middle) voxel representation, (right) mesh representation}
% \label{fig:3d_rep}
% \end{figure}

%\subsubsection{Voxels}
\vspace{1mm}
\noindent \textbf{Voxels.}
A voxel representation provides information on a regular grid in 3D space. The voxel (volume element) is analogous to pixels (pictures/pix elements) on which information of 2D images is placed. 
The information provided at every voxel includes occupancy, color, or other features.
A voxel representation can be obtained from a point cloud through the process of voxelization, which groups all features of 3D points within a voxel for later processing. 
The structured nature of 3D voxels allows processing such information similar to 2D methods, e.g., convolutions. 
In a 3D convolution, the kernel slides in three dimensions as opposed to two dimensions in 2D convolutions. 
On the other hand, a voxel representation is usually sparse as it contains a lot of empty volumes corresponding to space around objects. 
Additionally, since most capturing sensors collect information of object surfaces, object internals are also represented by empty volumes. 

%\subsubsection{Meshes}
\vspace{1mm}
\noindent \textbf{Meshes.}
A mesh is a collection of vertices, edges, and faces (polygons). 
The elementary component is the polygon, which is a planar shape defined by connecting a group of 3D vertices. 
Compared to the point cloud which only provides vertices locations, a mesh includes information of the object surface. 
The mesh is commonly used in computer graphics applications to represent 3D models. 
Nonetheless, processing the surface information directly using deep learning methods is not straightforward, and many techniques resort to sampling points from the surfaces in order to transform the mesh representation into a point cloud. 

\vspacesection
\subsection{3D Processing}
For 2D image understanding, the common representation on the regular grid is utilized. 
On the other hand, 3D data can be represented differently, and various methods have been proposed. 
Existing 3D deep learning approaches can be categorized as: (1) point-based, (2) voxel-based, and (3) projection-based.

\begin{figure}[!t]
\renewcommand{\arraystretch}{1.3}
\centering
\begin{tabular}{c}
\vspace{-2mm}\includegraphics[width=1\linewidth]{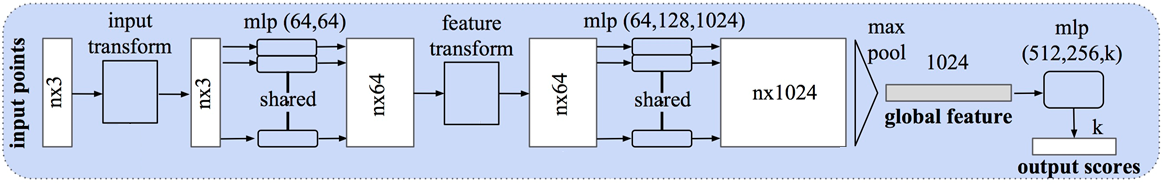} \\
\footnotesize Point-based method: Pointnet \cite{qi2017pointnet} (\copyright 2017 IEEE)\\
\vspace{-2mm}\includegraphics[width=1\linewidth]{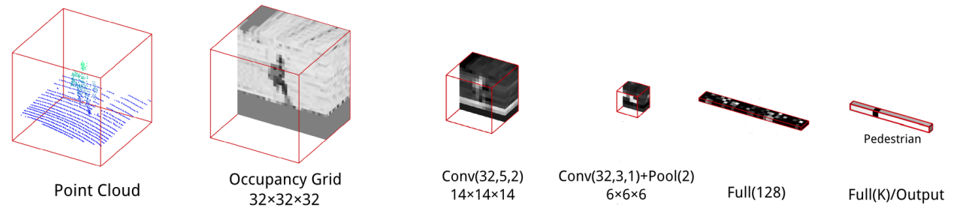} \\
\footnotesize Voxel-based method: Voxnet \cite{maturana2015voxnet} (\copyright2015 IEEE)\\
\end{tabular}
\caption{Comparison between voxel-based and point-based methods. A point-based method directly processes the point cloud, whereas a voxel-based scheme handles the data on a regular grid via voxelization. Figures from \cite{qi2017pointnet}  and \cite{maturana2015voxnet}.}
\label{fig:processing}
\vspacefigtext
\end{figure}

%\subsubsection{Point-Based Deep Networks}
\noindent \textbf{Point-Based Deep Networks.}
%A point cloud is a set of 3D points. 
%
Point-based methods directly process such data without transforming it into a regular set. 
Therefore, such methods extract feature information using permutation invariant techniques. 
PointNet \cite{qi2017pointnet} uses pointwise MLPs with the global max-pooling operator to extract features while being permutation invariant (Fig. \ref{fig:processing}). 
Nonetheless, PointNet does not capture local structures in the physical space around 3D points. 
As such, PointNet++ \cite{qi2017pointnet++} is proposed to combine local features at multiple scales. 
%This allows learning of local geometric layout.

A few recent approaches use graph neural networks (GNNs) to process point clouds. 
The nodes on the graph correspond to 3D points and information is passed through edges connecting the nodes. 
Dynamic Graph CNN \cite{wang2019dynamic} exploits local geometric structures by constructing a local neighborhood graph using K-nearest neighbor (kNN). 

Another approach to directly process 3D point clouds is to use continuous convolution operations. 
For example, SpiderCNN \cite{xu2018spidercnn} uses a family of polynomial functions as kernels for convolution. 
The kernel weights at neighboring points are thus dependent on the distance to those points. 
KPConv \cite{thomas2019kpconv} introduces a point convolution in which the kernel is represented as a set of points in the Euclidean space with kernel weights. 
On the other hand, PointConv \cite{wu2019pointconv} uses nonlinear functions of local 3D coordinates as convolution kernels with weight and density functions. 
The weight functions are learned with MLPs, whereas the density functions are learned by kernel density estimation.

%\subsubsection{Voxel-Based Deep Networks}
\noindent \textbf{Voxel-Based Deep Networks.}
Instead of processing unordered and irregular point cloud sets, numerous methods transform the 3D data into a regular grid by voxelization. 
In~\cite{maturana2015voxnet} 3D convolutions are applied to  the dense voxel grid for object recognition. 
Nevertheless, when compared to 2D images, the added dimension leads to a significant increase in the size of the data to process, leading to constraints on object size or voxel resolution. 
Furthermore, this approach is not computationally efficient as it does not benefit from the sparse nature of the 3D data. 
Alternatively, other methods \cite{graham20183d,choy20194d} operate convolutions at occupied voxels only, which greatly reduces the computational requirements. 
This allows processing at a higher resolution, which is reflected in a higher accuracy compared to the dense approach.
Other methods \cite{wang2017cnn,riegler2017octnet} propose to learn the 3D representation at higher resolutions by partitioning the space into an octree hierarchy. 
In the octree structure, densely occupied regions are modeled with high accuracy whereas empty regions are represented by large cells.% in the octree. 
Fig. \ref{fig:processing} shows a comparison between a voxel-based and a point-based method.

%\subsubsection{Projection-Based Deep Networks}
\noindent \textbf{Projection-Based Deep Networks.}
Another way to transform the irregular point cloud sets into a regular one is through projection. 
Once 3D data is projected onto a plane, 
numerous 2D methods can be used for analysis. 
Existing approaches include projecting point cloud sets into multiple views \cite{su2015multi,chen2017multi,kalogerakis20173d,kanezaki2018rotationnet}, onto a 2D plane for processing \cite{lang2019pointpillars}, or 
onto an estimated tangent plane and applying convolutions with continuous kernels  \cite{tatarchenko2018tangent}.

%can operate on 2.5D, which is the 3D projection

%can also project into other space (such as tangentconv)

\begin{figure}[t]
\centering
\includegraphics[width=\linewidth]{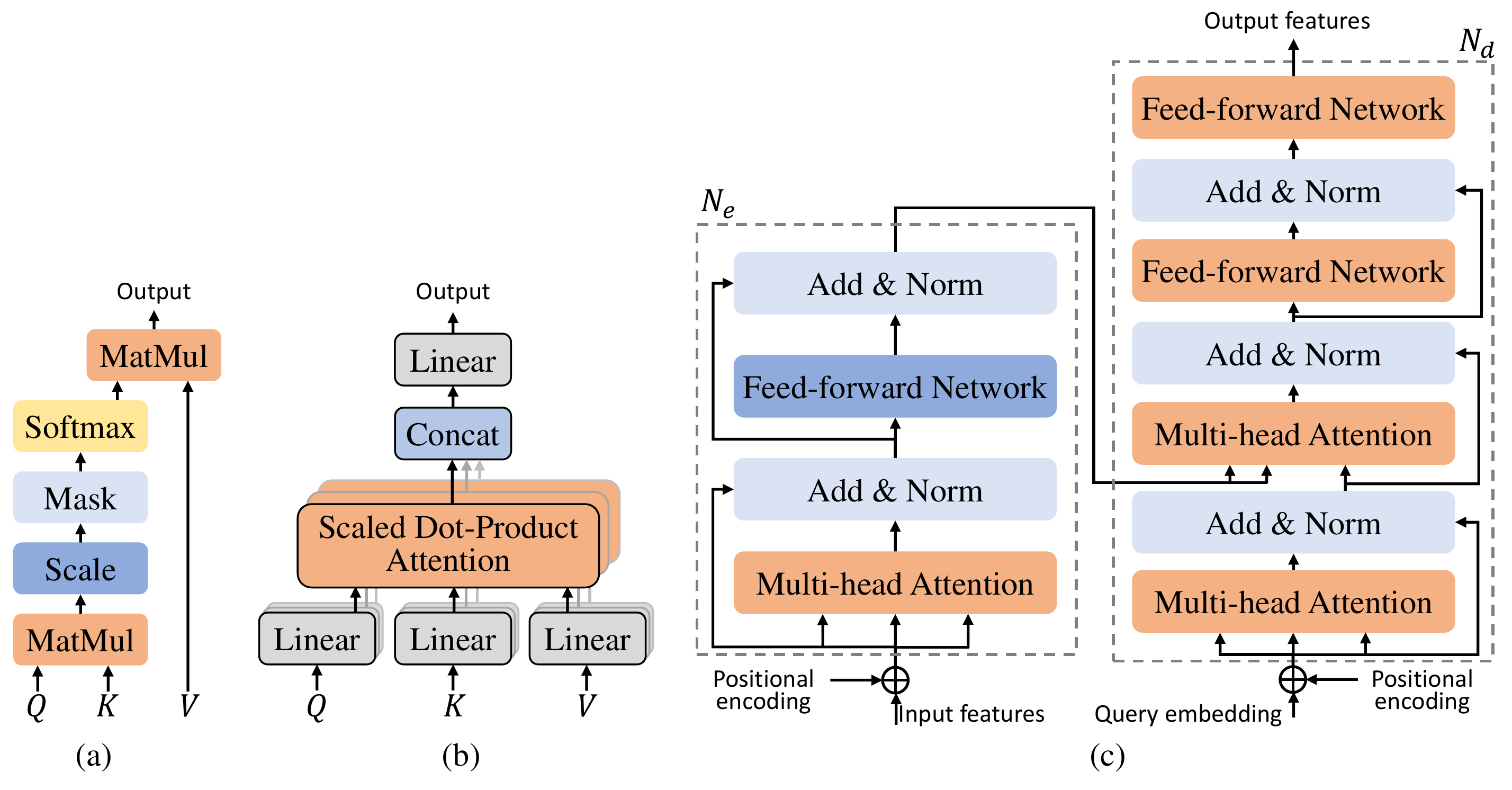}
\caption{Attention and transformer structures. (a) scale dot-product attention. (b) multi-head attention. (c) transformer. In both (a) and (b), there are three inputs: a query vector $Q$, a key vector $K$, and a value vector $V$, and a weighted output. In (c), the transformer consists of an encoder (left) and a decoder (right).}% The transformer usually consists of a transformer encoder and a transformer decoder.}
\label{fig:att_arch}
\vspacefigtext
\end{figure}

\vspacesection
\subsection{Transformer}

%\subsubsection{Self-Attention}
\vspace{1mm}
\noindent \textbf{Self-Attention.}
%Self-Attention and Multi-Head self-attention
Transformers \cite{Vaswani2017attention} have been widely used in numerous language and vision tasks. 
In transformers, scaled dot-product attention is the key, which aims to capture the dependencies between different input elements. 
Fig. \ref{fig:att_arch}(a) shows a typical scaled dot-product attention module. 
The attention module takes a query vector $Q$, a key vector $K$, and a value vector $V$ as inputs and generates a weighted sum of the values:
\begin{equation}
    Z=f_{att}(Q,K,V)=\text{Softmax}(\frac{QK^T}{\sqrt{d_k}})V,
    \label{eq:att}
\end{equation}
where $d_k$ represents feature dimension in the query and key that is used to scale the output of dot product operation. 

The scaled dot-product attention generates a single attention map to represent the relationship between the query and the key. 
To better represent the relationship, multi-head attention (Fig. \ref{fig:att_arch}(b)) aims to attend to information from different
sub-spaces. 
Multi-head attention consists of $h$  scaled dot-product attention modules and generates $h$ different outputs:
\begin{equation}
    Z_i=f_{att}(QW_i^q,KW_i^k,VW_i^v), i=1,...,h,
\end{equation}
where $f_{att}$ represents scaled dot-product attention in Eq. \ref{eq:att}, $W_i^q,W_i^k,W_i^v$ are learnable weight matrices for the query, key, and value. 
Finally, multi-head attention concatenates the $h$ outputs and feds it to a linear layer to generate the final output.

% \begin{figure}[t]
% \centering
% \includegraphics[width=0.6\linewidth]{images/figTransformerArch.jpg}
% \caption{Transformer architecture, which consists of an encoder (left) and a decoder (right). The transformer usually consists of a transformer encoder and a transformer decoder.}
% \label{fig:transformer}
% \vspacefigtext
% \end{figure}
% \subsection{Transformers}

%\subsubsection{Transformer Architecture}
\vspace{1mm}
\noindent \textbf{Transformer Architecture}
%Transformers usually adopt an encoder-decoder structure. 
%
Fig. \ref{fig:att_arch}(c) shows a typical transformer model based on an encoder-decoder structure. 
The encoder has $N_e$ identical blocks, where each block consists of a multi-head self-attention sub-layer and a feed-forward network. 
The multi-head self-attention sub-layer captures the relationship between different input elements, while the feed-forward network converts the features for each input element with multi-layer perceptrons. 
In addition, there is a residual connection and a normalization operation after each sub-layer. 
The decoder has $N_d$ identical blocks, where each block consists of a multi-head self-attention sub-layer, a multi-head cross-attention sub-layer, and a feed-forward network. 
The multi-head self-attention sub-layer captures the relationship between different decoder elements, while  the multi-head cross-attention sub-layer performs attention on the outputs of the encoder by taking the outputs of the encoder as the key and value. The feed-forward network converts the features for each input element with multi-layer perceptrons. 
Similar to the encoder, there is a residual connection and a normalization operation after each sub-layer in the decoder. 

%MH: you do not describe anything about feed-forward network or MLP. These are important parts of a transformer. 

%\subsubsection{Transformers Properties}
\vspace{1mm}
\noindent \textbf{Transformer Properties.}
The transformer properties are summarized as follows:
\begin{itemize}
    \item Transformers generate the outputs according to the relationship  between different elements. Namely, transformers can dynamically aggregate the inputs instead of learning static weights.
    %between the similarities
    % \item transformers can easily operate on irregular inputs, which is very suitable for 3D vision task.
    \item Transformers are permutation invariant. A common representation of 3D information is the point cloud data, which is an unordered set of points. 
    Therefore, a permutation invariant technique is required to ensure a consistent output for the same input (object).
    \item Transformers are capable of processing arbitrary-sized inputs. This is suitable for the 3D domain since the input data occurs in varying sizes.
    \item Transformers model long-range relationships. They are not bound to narrow receptive fields and are suitable for 3D vision tasks with scattered input. 
\end{itemize}

These properties show that using transformers in 3D computer vision tasks has good prospects.
As a result, transformer-based 3D vision tasks have attracted much attention in recent years.

\vspacesection
\section{Transformer Design in 3D Vision}\label{sec:applications}

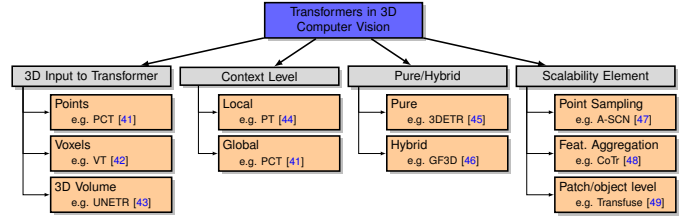
\begin{figure}[t]
\resizebox{\linewidth}{!}{
\begin{tikzpicture}[
  level 1/.style={sibling distance=45mm, very thick},
  edge from parent/.style={->,draw},
  >=latex]

% root of the the initial tree, level 1
\node[root] {Transformers in 3D Computer Vision}
% The first level, as children of the initial tree
  child {node[level 2] (c1) {3D Input to Transformer}}
  child {node[level 2] (c2) {Context Level}}
  child {node[level 2] (c3) {Pure/Hybrid}}
  child {node[level 2] (c4) {Scalability Element}};

% The second level, relatively positioned nodes
\begin{scope}[every node/.style={level 3}]
\node [below of = c1, xshift=15pt, yshift=1pt] (c11) {Points \\ \footnotesize ~~~ e.g. PCT \cite{guo2021pct}};
\node [below of = c11, yshift=-3pt] (c12) {Voxels \\ \footnotesize ~~~ e.g. VT \cite{mao2021voxel} };
\node [below of = c12, yshift=-3pt] (c13) {3D Volume \\ \footnotesize ~~~ e.g. UNETR \cite{hatamizadeh2022unetr}};

\node [below of = c2, xshift=15pt, yshift=1pt] (c21) {Local \\ \footnotesize ~~~ e.g. PT \cite{zhao2021point}};
\node [below of = c21, yshift=-3pt] (c22) {Global \\ \footnotesize ~~~ e.g. PCT \cite{guo2021pct}};
\node [below of = c3, xshift=15pt, yshift=1pt] (c31) {Pure \\ \footnotesize ~~~ e.g. 3DETR \cite{misra2021end}};
\node [below of = c31, yshift=-3pt] (c32) {Hybrid \\ \footnotesize ~~~ e.g. GF3D \cite{liu2021group}};

\node [below of = c4, xshift=15pt, yshift=1pt] (c41) {Point Sampling \\ \footnotesize ~~~ e.g. A-SCN \cite{xie2018attentional}};
\node [below of = c41, yshift=-3pt] (c42) {Feat. Aggregation \\ \footnotesize ~~~ e.g. CoTr \cite{xie2021cotr}};
\node [below of = c42, yshift=-3pt] (c43) {Patch/object level \\ \footnotesize ~~~ e.g. Transfuse \cite{zhang2021transfuse}};
\end{scope}

% lines from each level 1 node to every one of its "children"
\foreach \value in {1,2,3}
  \draw[->] (c1.189) |- (c1\value.west);

\foreach \value in {1,...,2}
  \draw[->] (c2.189) |- (c2\value.west);

\foreach \value in {1,2}
  \draw[->] (c3.189) |- (c3\value.west);
  
\foreach \value in {1,2,3}
  \draw[->] (c4.189) |- (c4\value.west);
  
\end{tikzpicture}}
\caption{Taxonomy of the transformer design in 3D Computer Vision. We group the methods into underlying approach differences pertaining to the input to the transformer, context level, its combination with other learning methods (pure/hybrid), and scalability element. We show the common choices in each group as well as example paper references. }
\label{fig:taxonomy}
\vspacefigtext
\end{figure}

The attention block captures long-range dependencies which facilitate learning context not fully exploited in convolution-based networks.
These long-range dependencies can play an important role in scene understanding especially when the local information is ambiguous.
Moreover, transformers can be applied to the sets, which is the natural representation of a point cloud. Unlike image representations, point clouds can occur in different lengths, sharing similarities to words in sentences. 
Given the success of transformers in NLP, one would hope that the transformer integration into the 3D domain would follow a similar trend. 
Additionally, transformers applied in 2D require adding position information to the feature information. 
In 3D, the position is available as coordinates of the points in the point cloud. 
The aforementioned properties of transformers have formed a ground for using the transformer architecture in the 3D domain. 
Nevertheless, there are numerous ways to integrate a transformer into a 3D application pipeline. 
Therefore, we discuss key characteristics of such integration in this section.
We base our discussion on the taxonomy shown in Fig. \ref{fig:taxonomy}.

% \subsubsection{Point transformer versus Voxel Transformer}
%\subsection*{3D Input to Transformer}
\vspace{1mm}
\noindent \textbf{3D Input to Transformer.}
Different 3D data representations can be processed using a transformer architecture. 
The choice of the data representation affects data size, data distribution, level of detail (granularity), and structure. 
Moreover, the data representation would allow transformer-based methods to be coupled with existing schemes for that specific representation.

The ability of a transformer to process unordered sets allows its application directly on point clouds.
%
%MH: the following sentences look good to me. 
%\textcolor{red}
The input to the transformer in this case would be the point coordinates, as well as any additional features used with non-transformer-based architectures, such as color, normal, and height from the ground, among many others. 
Since points belong to a continuous domain, an efficient sampling technique is required prior to processing.

Large point clouds can be cropped to a certain physical size to help process fewer points while keeping a fine resolution to capture local geometric features.
%
%MH: Euclidean space not euclidean space
A fixed size in the Euclidean space does not lead to a fixed number of points. 
Therefore, models that process fixed-sized inputs require sampling from the point cloud. 
Sampling can be achieved through random sampling, farthest point sampling, kNN sampling, or sampling from meshes, to name a few. 
Such sampling affects the number of points representing a given object since it depends on the scene complexity. %
While most methods do sampling in the pre-processing stages, some methods sample during the training process.%, eg kNN sampling. 
This can lead to a significant overhead in the training process.

An alternative to processing point clouds directly with a transformer is to convert the input to a regular grid. %
The equally spaced voxels allow similar representation of objects irrespective of the number of points in the point cloud. 
It also facilitates neighbor search, if required, since it can be searched with a hash table. 
On the other hand, a fine grid resolution is required to capture fine shape information, which leads to a cubic increase in the data to process. 
In addition, since 3D scenes are dominated by empty space, it is inefficient to process empty voxel grids. 
Processing occupied voxels would result in different sized inputs, and thus can be sampled similar to point clouds. Nonetheless, point density consistency and the easier search would still be present for the voxel representation.  

% \subsubsection{Local versus Global transformer}
%\subsection*{Context Level}
\vspace{1mm}
\noindent \textbf{Context Level.}
An efficient vision application should be able to capture fine local information as well as global context. 
In both cases, an increase in the computational requirement is encountered. Therefore, data is commonly processed on different scales to achieve both targets.

Transformers that process 3D information can be applied to a local neighborhood of points to capture local shape information. 
Similar to methods in the 2D domain, local pooling would allow processing on a different scale with a larger receptive field. 
The larger receptive field provides interaction between farther points for learning context.
% example point Transformer and pointformer

Since applying a transformer on local information requires a multi-layer application, a transformer that consumes the whole 3D data is also feasible. 
This reduces the necessity for local neighborhood sampling since the whole point cloud is used at once.
Nonetheless, the size of the input data is limited, and a good balance between point cloud coverage and density of points is needed.
% example 3detr

%\subsection*{Pure versus Hybrid Transformer}
\vspace{1mm}
\noindent \textbf{Pure and Hybrid Transformer.}
%\noindent \textbf{Pure Transformers}: 
Pure transformer architectures rely on attention layers to extract features and generate task specific output. 
In some cases, non-attention layers are utilized to encode the input or to supplement the attention layers. 
We consider an architecture as a pure transformer if it does not rely on previous non-attention architectures or backbones in the proposed pipeline.

%\noindent \textbf{Hybrid Architectures}: 
Since transformers are capable of capturing global context, they can be used to extract richer features.
One way to integrate transformers into deep learning architectures is the replace the feature extraction module with one that is attention/transformer based. %self-attention as an auxiliary module
Instead of fully relying on transformers to extract features, one can alternatively process local feature extraction using non-transformer-based methods, and then couple it with a transformer for global feature interaction.
Transformers can also be complemented by non-transformer layers to extract richer information. The complementarity can be caused by different resolutions to which each method can be applied.  

% \subsubsection{Relative versus Absolute Position Embedding ?}

%\subsection*{Scalability Element}
\vspace{1mm}
\noindent \textbf{Scalability.}
3D data incur more information when compared to 2D due to the additional third dimension. 
On the other hand, transformers are computationally expensive since they need to generate a large attention map, which has quadratic complexity with respect to the input size. 
Given the increased data size and transformer input size limit, it requires a sampling scheme to enable processing. 
Typical approaches decrease the input size to the transformer, and thus allowing scalability including: farthest point sampling for point clouds with kNN feature aggregation, voxel feature aggregation for low-resolution volumetric representation, patch feature embedding, feature downsampling using CNN-based methods, and 
object level self-attention. 

%\begin{itemize}
%  \item farthest point sampling for point clouds with kNN feature aggregation
%  \item voxel feature aggregation for low-resolution volumetric representation
%  \item patch feature embedding
%  \item feature downsampling using CNN-based methods
%  \item object level self-attention 
%\end{itemize}

\vspacesection
\section{Transformers in 3D Vision: Applications}
The transformer architecture has been integrated into various 3D vision applications. 
In this section, we review methods based on the targeted 3D vision tasks, including object classification, object detection, segmentation, point cloud completion, pose estimation, and others.

\vspacesection
\subsection{Object Classification}
%In this section, we review methods that classify objects given their 3D representation. 
%
We first give an overview of methods that employ the transformer within a defined local region is presented, and then discuss methods that apply the transformer on the global level. 
Table \ref{tab:classification} shows an overview of these methods using based on the above-defined taxonomy.

% \begin{table*}[!t]
% \renewcommand{\arraystretch}{1.3}
% \caption{Classification Methods}
% \label{table_example}
% \centering
% \begin{tabular}{|c|c|c|c|c|c|}
% \hline
% Method & Input to Transformer & Scalability ~~~~~~~ Element & Pure or hybrid & Type of attention & Context type\\
% \hline
% Three & Four & Four & Four & Four & Four\\
% \hline
% \end{tabular}
% \end{table*}

%MH: For all the bales, the gray colors do not really fill up each row (i.e., some empty space below the lines). Please try to fill the entire rows (no space below the lines). 
%jean: updated the colors between the table lines
\renewcommand{\arraystretch}{0.5}
\begin{table*}[t]
    \centering
    \tiny
    \caption{Overview of transformer-based methods for classification. Important attributes for the transformer integration are shown here, which include the input,  sampling element that enables transformer processing,  architecture (pure  or hybrid), and context level on which the transformer operates. A highlight of the main contributions is also included. All the methods also perform object part segmentation except the ones with an asterisk (*)}
    %\begin{tabular}{|p{0.15\textwidth}|p{0.05\textwidth}
    %|p{0.1\textwidth}|p{0.05\textwidth}
    %|p{0.1\textwidth}|p{0.4\textwidth}|}
    \begin{tabular}{p{0.11\textwidth}p{0.05\textwidth}
    p{0.1\textwidth}p{0.05\textwidth}
    p{0.07\textwidth}p{0.47\textwidth}}
    %\hline
    \toprule
    \belowrulesepcolor{gray!30!}
\rowcolor{gray!30!} Method & Input & Scalability Element & Architecture & Context &  Highlight \\ \aboverulesepcolor{gray!30!} \midrule
Point Transformer   \cite{zhao2021point} & points & farthest point sampling & pure & local & applies self-attention in a local neighborhood, Transition down   and up to increase receptive field \\ \midrule
Point Transformer   \cite{engel2021point} & points & local (ball query)/ global (FPS) & with own sortnet & local / global & extract ordered local feature sets from different subspaces   (SortNet), global and local-global attention \\ 
%\hline
\midrule
Attentional \t ShapeContextNet   \cite{xie2018attentional} & points & random sampling & pure & global & replace hand-designed bin partitioning and pooling by a   weighted sum aggregation function with input learned by self-attention \\ %\hline
\midrule
Yang \etal \cite{yang2019modeling} & points & random sampling & pure & global & absolute and relative position embedding as input to attention   module,  group attention similar to   depthwise separable convolutions \cite{chollet2017xception} and channel   shuffle \cite{zhang2018shufflenet}. \\ 
%\hline
\midrule
PCT \cite{guo2021pct} & points & farthest point sampling & pure & global & offset-attention calculates the element-wise difference between the self-attention and the input features \\ 
%\hline
\midrule
PVT \cite{zhang2021pvt} & voxels, points & local (hash table), global (all) & pure & local / global & combines voxel-based and point-based transformer models to   extract feature information \\ 
%\hline
\midrule
TransPCNet* \cite{zhou2022sewer} & points & kNN aggregation & hybrid & global & feature embedding module and attention module to learn   features to detect defects in sewer represented by 3D point clouds \\ 
%\hline
\midrule
Adaptive Wavelet \t Transformer   \cite{huang2021adaptive} & points from graph & k nearest neighbor   & hybrid & global & perform multi-scale analysis to generate visual representation decomposition using the lifting scheme approach \\ 
%\hline
\midrule
3DCTN* \cite{lu20223dctn} & points & query ball & hybrid & local & combines graph convolution layers(local feature aggregation)   with transformers (global feature learning ) \\ %\hline
\midrule
DTNet \cite{han2021dual} & points & farthest point sampling & pure & global & Dual Point Cloud Transformer module to capture long-range   position and channel correlations \\ 
%\hline
\midrule
CpT \cite{kaul2021cpt} & points & k nearest neighbor & pure & local / global & uses a dynamic point cloud graph to create a point embedding   that is fed into the transformer layer \\ %\hline
\midrule
LFT-Net \cite{gao2022lft} & points & k nearest neighbor & pure & local & local feature transformer with local position encoding,   self-attention pooling function for feature aggregation \\ 
%\hline
\midrule
Point-BERT   \cite{yu2021point} & points & FPS and point patches & hybrid & global & point tokenization which converts a point cloud into discrete   point tokens, and masked point modeling for pre-training \\ 
%\hline
\midrule
Liu \etal   \cite{liu2022group} & points & FPS, kNN & pure & local / global & radius-based feature   abstraction for better feature extraction, group-in-group relation-based   transformer architecture \\ 
%\hline
\midrule
Pang \etal   \cite{pang2022masked} & points & FPS, kNN (or FPS and point patches) & pure & global & Transformer-based autoencoder with asymmetric design and   shifting mask tokens operation for pre-training \\ 
%\hline
\midrule
PAT   \cite{cheng2021patchformer} & voxels & voxelization & hybrid & local / global & patch attention module (PAT) and a multi-scale attention module (MST) \\ 
%\hline
\midrule
3CROSSNet   \cite{han20223crossnet} & points & farthest point sampling & hybrid & local / global & Point-wise Feature Pyramid, Cross-Level Cross-Attention, and CrossScale Cross-Attention \\ 
%\hline
\midrule
3DMedPT \cite{yu20213d} & points & farthest point sampling & hybrid & global & local context augmentation, relative positional embedding, and   local context aggregation at query \\ 
%\hline
\midrule
Wu \etal*   \cite{wu2021centroid} & points \t \&others & soft k-means clustering & pure & global & centroid attention: summarizing self-attention feature mapping to a smaller number of outputs \\ 
%\hline
\midrule
MLMSPT   \cite{han2021point} & points & farthest point sampling & pure & local / global & point pyramid transformer followed by multi-level and   multi-scale transformer \\ 
%\hline              
\bottomrule
    \end{tabular}
    \label{tab:classification}
    \vspacefigtext
\end{table*}

\begin{figure}[t]
\renewcommand{\arraystretch}{0.5}
% \label{table_example}
\centering
\begin{tabular}{cl}
\tiny \rotatebox{90}{\makecell{Point Trans. \cite{zhao2021point} \\ \copyright 2021 IEEE}} & \includegraphics[width=.88\linewidth]{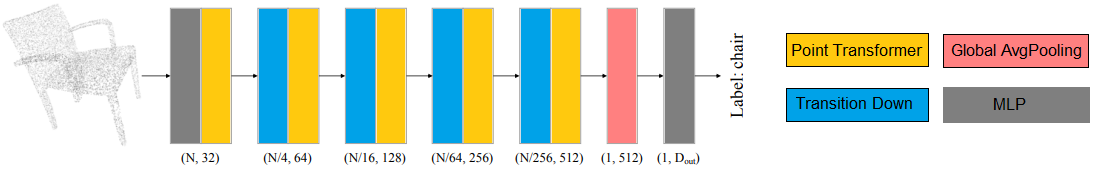} \\
\tiny \rotatebox{90}{\hspace{5mm}  PCT \cite{guo2021pct}} &
\includegraphics[width=.88\linewidth]{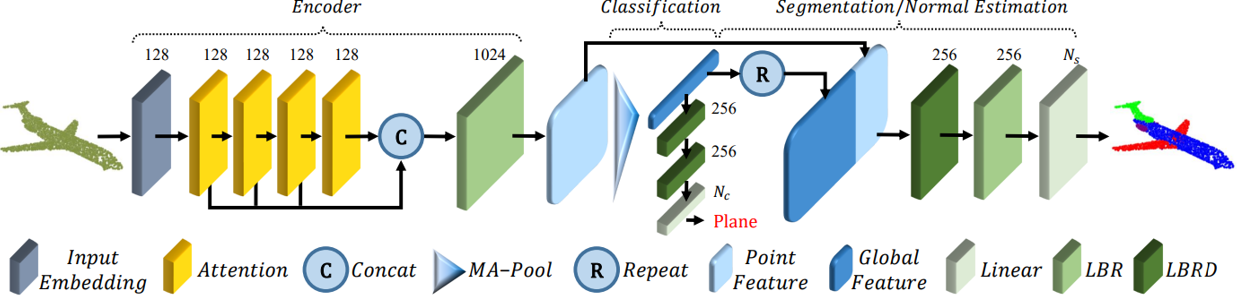} \\
\end{tabular}
\caption{Example of local (Point Transformer) and global (PCT) methods. The local method uses multiple transformer layers at different scales to achieve a larger receptive field. Global methods apply multiple attention layers for richer feature representation. Figure from \cite{zhao2021point} and \cite{guo2021pct}.}
\label{fig:classification_example}
\vspacefigtext
\end{figure}

%\subsubsection*{Local Transformers}
\vspace{1mm}
\noindent \textit{Local Transformers.} 
Point Transformer \cite{zhao2021point} applies self-attention in the local neighborhood of each data point. 
A point transformer block consists of the attention layer, linear projections, and residual connection (Fig. \ref{fig:classification_example}). 
Additionally, instead of using the 3D point coordinates as position encoding, an encoding function is used with linear layers and ReLU nonlinearity. 
To increase the receptive field of the proposed transformer architecture, transition down layers are introduced, as well as transition up to retrieve the original data size. %Point2Sequence \cite{liu2019point2sequence} uses attention to capture fine-grained contextual information of local point cloud regions. It first samples local regions, then an MLP is used to extract features for each region. An attention mechanism is then employed to learn correlation between different areas in the local region.

3DCTN \cite{lu20223dctn} proposes to combine graph convolution layers with transformers. 
The former learns local features efficiently, whereas the latter is capable of learning global context. 
Taking the point cloud with normals as input, the network consists of two modules that downsample the point set, with each module having two blocks: the first block is a local feature aggregation module using a graph convolution, and the second block is a global feature learning module using a transformer consisting of offset-attention and vector attention. 
LFT-Net \cite{gao2022lft} proposes a local feature transformer network that uses self-attention to learn features of point clouds. 
It also introduces a Trans-pooling layer that aggregates local features to reduce the feature size.

%\subsubsection*{Global Transformers}
\vspace{1mm}
\noindent \textit{Global Transformers.}
%MH: it would be good to add one or two sentences in each paragraph to describe the methods discussed in each paragraph (i.e., underlying common themes/approaches of these methods --- why you cluster them in each paragraph). Otherwise, it reads likes a laundry list. 
%jean: updated the global transformers paragraph
At the global scale, the attention module has been integrated into various parts of the network with different inputs and position embeddings.
Attentional ShapeContextNet \cite{xie2018attentional} is one of the early adoptions of self-attention for point cloud recognition.  
To learn shape context, the self-attention module is used to select contextual regions, aggregate and transform features.
This is carried out by replacing hand-designed bin partitioning and pooling with a weighted sum aggregation function with input learned by self-attention applied on all the data. 
In \cite{huang2021adaptive}, Adaptive Wavelet Transformer first performs multi-resolution analysis within the neural networks to generate visual representation decomposition using the lifting scheme technique. 
The generated approximation and detail components capture geometric information that is of interest for downstream tasks. %
A transformer is then used to pay different attention to features from approximation and detail components and to fuse them with the original input shape features. 
TransPCNet \cite{zhou2022sewer} aggregates features using a feature embedding module, feeds them into separable convolution layers with a kernel size of 1, and then uses an attention module to learn features to detect defects in sewers represented by 3D point clouds.

Other methods propose variations of the attention module.
Yang \etal \cite{yang2019modeling} develop Point Attention Transformers (PATs) by applying an attention module to a point cloud represented by both absolute and relative position embeddings. 
The attention module uses a multi-head attention design with group attention, which is similar to depthwise separable convolution \cite{chollet2017xception}, in addition to channel shuffle \cite{zhang2018shufflenet}. 
Point Cloud Transformer (PCT) \cite{guo2021pct} applies offset-attention to the input point embedding. 
The offset-attention layer calculates the element-wise difference between the self-attention features and the input features (Fig. \ref{fig:classification_example}). 
It also uses a neighbor embedding by sampling and grouping neighboring points for better local feature representation. 
DTNet \cite{han2021dual} aggregates point-wise and channel-wise multi-head self-attention models to learn contextual dependencies from the position and channel. 

Some methods focus on pre-training the transformer by masking parts of the input.
Point-BERT \cite{yu2021point} first partitions the input point cloud into point patches, inspired by Vision Transformers \cite{dosovitskiy2020image}, and uses a mini-Pointnet \cite{qi2017pointnet} to generate a sequence of point embeddings. 
The point embeddings are then used as an input to a transformer encoder, which is pre-trained by masking some point embeddings with a mask token, similar to \cite{devlin2018bert}. 
The tokens are obtained using a pre-learned point Tokenizer that converts the point embeddings into discrete point tokens. 
Similarly, Pang \etal \cite{pang2022masked} divide the input point cloud into patches and randomly mask them during pre-training. 
A transformer-based autoencoder is used to retrieve the masked point patches by learning high-level latent information from unmasked point patches.

%\subsubsection*{Local and Global Transformers}
\vspace{1mm}
\noindent \textit{Local and Global Transformers.}
%MH: as mentioned above, it is best to use one or two sentences to give an overview of the underlying theme of the methods described in each paragraph (why they are discussed in the same paragraph?).
% not grouped yet
%jean: updated the local and global transformers paragraph
%Researchers made use of the transformer architecture to learn local and global information.
Numerous methods have been proposed to use transformer architecture for learning both local and global information. 
For that, the transformer has been deployed at various stages to process different information.
Engel \etal \cite{engel2021point} employ local-global attention to capture local and global geometric relations and shape information.
The input features to the attention module are local features of an ordered subset, which is learned via a permutation invariant network module. 
In \cite{kaul2021cpt}, CpT uses a dynamic point cloud graph to create a point embedding that is fed into the transformer layer.
The transformer layer is comprised of sample-wise attention that dynamically processes local point set neighbors as well as inter-point attention.
On the other hand, Liu \etal \cite{liu2022group} group points using farthest distance sampling with K nearest neighbors. 
It then uses group abstraction and radius-based feature abstraction to obtain group features. 
A transformer is then used with groups as well as across all point groups.
3DMedPT \cite{yu20213d} embeds local point cloud context by downsampling the points and then grouping local features similar to DGCNN \cite{wang2019dynamic}. 
It proposes the use of relative positional embeddings and local response aggregation at query. 

Point-Voxel Transformer (PVT) \cite{zhang2021pvt} combines voxel-based and point-based Transformer models to extract feature information. 
The voxel-based model captures local features efficiently because of regular data locality. 
It uses a local attention module, whose computational complexity is linear with respect to the input voxel size. 
The point-based model captures global features and also remedies the information loss during voxelization. 
It uses  relative attention which is a self-attention variant that considers pairwise relationships or distance between input points. 

On the other hand, some methods focus on applying attention on multiple scales.
%
%MH: it is confusing to say PAT method consists of PAT and MST. Revise it. 
%PAT \cite{cheng2021patchformer} constructs a voxel-based architecture that integrates a patch attention module (PAT) and a multi-scale attention module (MST). 
Patchformer \cite{cheng2021patchformer} constructs a voxel-based architecture that integrates a patch attention module (PAT) and a multi-scale attention module (MST).
The PAT module applies weighted summation over a small set of bases to capture the global shape, and thus achieves linear complexity to the input size whereas the MST module applies attention to features of different scales.
In \cite{han20223crossnet},  3CROSSNet first extracts multi-scale features using a point-wise feature pyramid module.
Cross-attention is then applied across levels to learn inter-level and intra-level dependencies. 
Another cross-attention module is applied across scales to better represent between scales and within scales interactions.
MLMSPT \cite{han2021point} proposes a point pyramid transformer that captures features from multiple levels and scales. 
A multi-level transformer and a multi-scale transformer are then used to capture contextual information from different levels and scales. 

Wu \etal \cite{wu2021centroid} introduce centroid attention, in which self-attention maps information in the inputs into a smaller output. 
During training, a soft K-means clustering objective function is optimized. 
%which allows clustering of information based on similarity. 
%
The centroid attention then transforms the input sequence into the set of centroids.

\begin{table*}[t]
\renewcommand{\arraystretch}{0.5}
    \centering
    \tiny
    \caption{Overview of 3D object detection methods using the transformer architecture. A highlight of the main design attributes and contributions is shown here. These methods use a variety of input representations, employ multiple sampling strategies for scalability, use a pure or hybrid transformer integration, and apply the transformer locally or globally.}
     %   \begin{tabular}{|p{0.12\textwidth}|p{0.05\textwidth}
    %|p{0.12\textwidth}|p{0.05\textwidth}
    %|p{0.07\textwidth}|p{0.42\textwidth}|}
    \begin{tabular}{p{0.12\textwidth}p{0.05\textwidth}
    p{0.12\textwidth}p{0.05\textwidth}
    p{0.06\textwidth}p{0.44\textwidth}}
    \toprule
    \belowrulesepcolor{gray!30!}
\rowcolor{gray!30!} Method & Input & Scalability Element & Architecture & Context & Highlight \\ \aboverulesepcolor{gray!30!} \midrule
Pointformer \cite{pan20213d} & points & Linformer for scalability & pure & global/ local & a feature learning block with a local, local-global, and   global transformer \\ \midrule
Voxel Transformer \cite{mao2021voxel} & voxels & voxel discretization & pure & local/ dilated & multi-head self-attention on non-empty voxels through local   attention and dilated attention \\ \midrule
Sheng \etal \cite{sheng2021improving} & points & proposal to point attention & hybrid & global & uses raw points and proposals as input into a channel-wise   transformer with a proposal-to-point encoding module and a channel-wise   decoding module \\ \midrule
Liu \etal \cite{liu2021group} & points & k-closest points sampling & hybrid & global & stacked multi-head self-attention and multi-head   cross-attention to extract and refine object representations for object   candidates \\ \midrule
DETR3D \cite{wang2022detr3d} & features & object queries & hybrid & - & multi-head attention to refine object queries by incorporating   object interactions, similar to DETR \cite{carion2020end}. \\ \midrule
3DETR \cite{misra2021end} & points & pointnet++ aggregation & almost pure & global & A transformer encoder is applied directly on the point cloud for   extracting feature information, and a transformer decoder to predict 3D   bounding boxes \\ \midrule
SA-Det3D \cite{bhattacharyya2021sa} & points, voxels, pillars & attend to salient regions & hybrid & global & augment multiple convolution-based methods with full   self-attention or deformable self-attention \\ \midrule
M3DETR \cite{guan2022m3detr} & points, voxels & applied on output & hybrid & global & combines raw points, voxels, and bird-eye view representations   under a unified transformer-based architecture \\ \midrule
Fan \etal \cite{fan2021embracing} & voxels & regional grouping & pure & local & transformer operates with sparse regional attention on   voxelized input \\ \midrule
Fast Point Transformer \cite{park2021fast} & voxels & (centroid aware voxelization) & pure & local & speed-up local self-attention networks with voxel hashing   architecture and centroid-aware voxelization and devoxelization \\ \midrule
Voxel Set Transformer \cite{he2022voxel} & voxels & voxelization & hybrid & local & A voxel-based set attention module with two cross-attentions. It applies self-attention to token clusters with varying sizes and processes them with a linear complexity. \\ \midrule
ARM3D \cite{lan2022arm3d} & points & FPS & hybrid & global & attention module to learn relation features for proposals \\ \midrule
Yuan \etal \cite{yuan2021temporal} & points & only temporal channel & hybrid & temporal & temporal encoder and    spatial decoder with a multi-head attention mechanism to aggregate   information from the adjacent video frames \\ \midrule
Dao \etal \cite{dao2022attention} & voxels & voxel discretization & hybrid & global & vector attention that learns different weights for the point feature channels\\ \midrule
MLCVNet \cite{xie2020mlcvnet} & points &  pointnet++ aggregation & hybrid & global & builds on top of \cite{qi2019deep} and uses self-attention  modules to learn contextual information at the patch, object, and global scene levels \\ \midrule
Yin \etal \cite{yin2020lidar} & pillars & discretized pillar nodes & hybrid & local & spatial transformer attention and temporal transformer   attention on point pillars \\ \midrule
SCANet \cite{lu2019scanet} & BEV & after encoder & hybrid & global & attention after VGG encoder for point cloud BEV and RGB \\ \midrule
MonoDETR \cite{zhang2022monodetr} & images & feature downsampling & hybrid & global & predicts and encodes depth to use it as input into a   depth-aware decoder \\ \midrule
Transfusion \cite{bai2022transfusion} & BEV, images & feature map from conv backbone & hybrid & global & convolution backbone to extract feature maps, transformer decoder to fuse LiDAR based queries with image features \\ \midrule
CAT-Det \cite{zhang2022cat} & points, ~~~images & ball query & hybrid & local/ global & combines a Pointformer applied on point cloud with Imageformer   applied on RGB images \\ \midrule
PETR \cite{liu2022petr} & images & conv-based encoder & hybrid & global & fuses 3D features from multi-view images with 2D features \\ \midrule
BoxeR \cite{nguyen2021boxer} & BEV, images & conv encoder for 2D, PointPillar for 3D & hybrid & local & applies attention to a sampled grid within a box in 2D and 3D \\ \midrule
BrT \cite{wang2022bridged} & points, ~~~images & pointnet++ aggregation & pure & global & bridging point tokens (from point cloud) and patch tokens from   images, point-to-patch projection \\ \midrule
VISTA \cite{deng2022vista} & BEV, RV & voxelization, projection to BEV and RV & hybrid & global & replace MLP in attention with convolutions, apply on BEV and   RV \\ \midrule
PDV \cite{hu2022point} & voxels & voxel discretization & hybrid & global & voxel centroid localization, density-aware RoI grid pooling,   grid point Self-Attention \\ \bottomrule
\end{tabular}
\label{tab:detection}
\vspacefigtext
\end{table*}

\vspacesection
\subsection{3D Object Detection}
Numerous attention-based methods have been proposed for 3D object detection. 
Table \ref{tab:detection} shows an overview of these methods and the adopted transformers. 
Most of these methods are applied to one domain: indoors or outdoors. 
Limiting applications to one domain is due to the varied modality of data collected indoors compared to outdoors, where RGB-D sensors are the common indoor 3D sensors and LiDAR is common outdoors. 
This leads to different dataset distributions, densities, and ranges. 
Nevertheless, methods from one domain can be applied to another domain having the same representation, but often require substantial adaptation to achieve competitive results. 

One of the methods that have been applied indoors and outdoors is Pointformer \cite{pan20213d}. 
It uses a transformer-based feature learning block with three parts: a local transformer to capture fine information in the local region, a local-global transformer to integrate the learned local features with the global information, and a global transformer to capture the global context. 
The local transformer aggregates features from the local regions into a subsampled set of points, thus reducing the computational requirements. 
Other methods have been only applied to either indoor or outdoor datasets.

\begin{figure*}[t]
\renewcommand{\arraystretch}{0.5}
% \label{table_example}
\centering
\begin{tabular}{c}
\includegraphics[width=\linewidth]{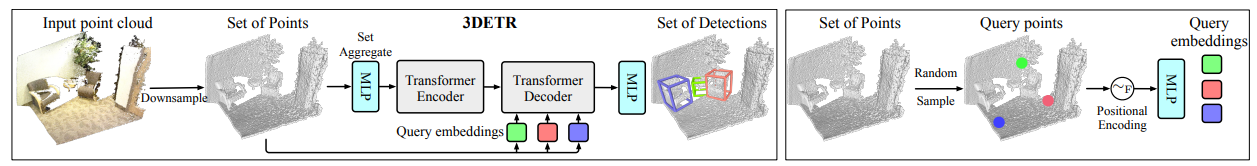} \\
\end{tabular}
\caption{3DTER: Pure transformer-based architecture for indoor 3D object detection. It consists of a transformer encoder applied directly on the point cloud for extracting feature information, and a transformer decoder to predict 3D bounding boxes. The decoder transformer is similar to DETR \cite{carion2020end} with adaptation to 3D detection through non-parametric query embeddings and Fourier positional embeddings. Figure from \cite{misra2021end} (\copyright 2021 IEEE).}
\label{3detr}
\vspacefigtext
\end{figure*}

%\subsubsection*{Indoor Scene Object Detection}
\vspace{1mm}
\noindent \textit{Indoor Object Detection}
 MLCVNet \cite{xie2020mlcvnet} builds on top of \cite{qi2019deep} and uses self-attention modules to aggregate contextual information at multiple levels, namely patch, object, and global scene levels. 
 At the patch level, the attention module is used to generate better voting to object centroid points. 
 The object level attention module captures contextual information among proposals, and 
 the global level attention module uses patch and cluster information to learn global scene context. 
 3DETR \cite{misra2021end} proposes an end-to-end transformer consisting of two modules: a transformer encoder applied directly on the point cloud for extracting feature information, and a transformer decoder to predict 3D bounding boxes (see Fig. \ref{3detr}). 
 The decoder transformer layers use non-parametric query embeddings from seed points for better 3D detection. 

 Liu \etal \cite{liu2021group} use a transformer module to extract and refine object representations from object candidates with features learned using PointNet++ \cite{qi2017pointnet++}. 
 The transformer module consists of multiple multi-head self-attention and multi-head cross-attention and operates on a subset of the points, sampled using the k-closest point sampling technique. 
 ARM3D \cite{lan2022arm3d} uses an attention-based module to extract fine-grained relations among proposal features learned using non-transformer-based architectures. 
 The objectness score is used to choose the proposals, and each proposal is matched with other proposals to learn relation contexts.

In \cite{wang2022bridged}, BrT uses a transformer to allow interactive learning between images and point clouds. 
It adopts conditional object queries of aligned points and image patches and adds point-to-patch projection for better learning.

%MH: similar to early section. It is best to add one or two sentences to give an overview of the methods discussed in the same paragraph. Otherwise, it reads like a long laundry list (especially from MonoDETR, Transfusion.... to the end of this subsection). 
%jean: updated the outdoor scenes object detection section
%\subsubsection*{Outdoor Object Detection}
\vspace{1mm}
\noindent \textit{Outdoor Object Detection.}
For outdoor environments, the transformer architecture has been utilized to process data from different sources and in different representations. 
Numerous transformer models with voxel representations have been proposed.
Among these methods is the Voxel Transformer \cite{mao2021voxel}, which processes an input voxel grid through a series of sparse and submanifold voxel modules. 
It performs multi-head self-attention on non-empty voxels through local attention and dilated attention. 
A voxel query mechanism is used to accelerate searching for non-empty voxels, benefitting from having the data on a regular grid.
%MH: do not say "claim"... bad connotation. 
%Fan \etal \cite{fan2021embracing} claim that downsampling the feature maps in 3D similar to the 2D domain would lead to a loss of  information. 
Fan \etal \cite{fan2021embracing} show that downsampling the feature maps in 3D similar to the 2D domain would lead to a loss of  information, and propose a single stride transformer to maintain the same resolution throughout the whole network. 
It also uses a voxelized input, but the transformer operates with sparse regional attention to reduce the computational requirement for the transformer modules. 
Fast Point Transformer \cite{park2021fast} is developed to speed up local self-attention networks. 
Since local self-attention usually requires finding the k-nearest points, it is usually a bottleneck.
The proposed self-attention module learns on a point cloud with voxel hashing architecture, which allows fast neighborhood selection and is coupled with a centroid-aware voxelization and devoxelization to embed continuous 3D coordinates. 
Recently, Voxel Set Transformer \cite{he2022voxel} presents a global approach to model long-range dependencies in a point cloud. 
It introduces a voxel-based set attention (VSA) module, which consists of two cross-attentions as a replacement for the self-attention, and can process inputs of varying sizes in parallel with linear complexity.
PDV	\cite{hu2022point} uses 3D sparse convolutions to extract feature information from a voxelized 3D scene, followed by a region proposal network head to generate bounding boxes. 
Voxel features are then pooled and used as an input to a self-attention module to refine the bounding boxes.

Other methods choose to apply transformers with a point cloud representation.
In \cite{sheng2021improving}, Sheng \etal~supplement a two-stage 3D detector with self-attention modules. 
This method first generates proposals via a 3D voxel-based region proposal network and then uses the raw points and the proposals as input into a channel-wise transformer in order to enrich the proposals with global context information. 
The channel-wise transformer consists of a proposal-to-point encoding module and a channel-wise decoding module that transforms the encoded features into final object proposals comprising confidence prediction and box regression.
On the other hand, PLNL-3DSSD \cite{liu2021plnl} uses local and non-local attention with set abstraction modules to model relationships between objects. 
%MH: this is an example where you could improve. In many paragraphs, you simply state what has been done using a few sentences. However, a good review should group several methods together and describe the underlying themes and differences. In addition, you can also discuss the strength and weakness with respect to other methods (to have a broader view). There is no sufficient depth to show your authority view of this topic and the review would only serve as a summary of existing methods. In most paragraphs of this "Indoor and outdoor scene" section, you almost always start with a method XYZ presents/proposes... and then a few sentences to describe XYZ. It reads very mechanical/boring. It is best to spice it up with variations. In some cases, Smith et al present... In some cases, In [10], a model based on... In some cases, "On the other hand, XYZ presents...". In some cases, "Recently, XYZ is developed to ..."...

Moreover, numerous methods apply the transformer to multi-view images or BEV, benefitting from the advancement of transformer application to images.
Transfusion \cite{bai2022transfusion} uses convolution backbones to extract LiDAR BEV feature map along with an image feature map. 
A transformer-based decoder takes object queries as input and outputs initial bounding box predictions using the LiDAR information. 
Next, a spatially modulated cross attention mechanism then performs fusion between camera image features and LiDAR object queries.
SCANet \cite{lu2019scanet} extracts features from RGB images and point cloud bird-eye-view using two VGG-16 encoders. 
A spatial-channel attention module is then employed to extract multi-scale and global context features to recalibrate the features.  
BoxeR \cite{nguyen2021boxer} introduces box attention, which learns attention weights for points sampled on a grid within boxes. 
In 2D, it uses convolution encoder features for proposals as input  and generates object queries. 
The object queries are then decoded into bounding boxes using instance attention. 
It learns attention weights that are invariant to rotation, so another transformer is used on the bird-eye's view to generate 3D bounding boxes.
Recently, MonoDETR \cite{zhang2022monodetr} modifies DETR to generate 3D bounding boxes from monocular images. 
The modification includes adding depth features to the input of the transformer. 
The depth features are generated using a depth predictor and a depth encoder. The transformer consists of a depth-aware decoder with self-attention as well as visual and depth cross-attention.
VISTA \cite{deng2022vista} proposes to replace the linear projections in the regular attention module with convolutional operators.
It applies the proposed attention to the projection of the features of a voxelized 3D scene into two views, bird-eye-view and range view. 

For point cloud videos, Yuan \etal \cite{yuan2021temporal} present a Temporal-Channel Encoder and a Spatial Decoder for 3D LiDAR-based video object detection.
The Temporal-Channel encoder is used to learn relations between different frames utilizing multi-head attention mechanism. 
The spatial decoder also utilizes a multi-head attention mechanism to aggregate relevant information of the adjacent video frames.

Since different representations might provide complementary information, some works have used the transformer with multiple representations.
SA-Det3D \cite{bhattacharyya2021sa} proposes to augment multiple convolution-based methods operating on points, voxels, and pillars with self-attention modules. 
It introduces two variants for self-attention: a full self-attention module, which is a pairwise self-attention mechanism, and a deformable self-attention module that learns deformations over randomly sampled locations to cover the most representative and informative parts.
On the other hand, M3DETR \cite{guan2022m3detr} aggregates information from raw points, voxels, and bird-eye view, under a unified transformer-based architecture. 
The transformers enable interactions among multi-representation, multi-scale, multi-location feature attention. Dao \etal \cite{dao2022attention} propose to use vector attention to refine voxel-based Region Proposal Networks. 
Compared to the multi-head attention, vector attention learns different weights for the different point feature channels, therefore it enables to capture richer information into the Regions of Interest and pooled points. 
CAT-Det \cite{zhang2022cat} combines a Pointformer applied on a point cloud with Imageformer applied on RGB images. 
The two modalities are then complemented with cross-modal feature interaction and multi-modal feature aggregation using a transformer.% architecture.

On the other hand, Yin \etal \cite{yin2020lidar} use spatial features to encode information on a point cloud discretized by pillars.
The spatial features are extracted from a given point cloud using graph-based operations and 2D CNN. 
Features from consecutive frames are then passed to a spatio-temporal transformer module that is constituted of spatial transformer attention as well as Temporal transformer attention.

Other methods use the transformer to fuse or refine information generated by non-transformer methods.
DETR3D \cite{wang2022detr3d} uses multi-view RGB images to detect objects in 3D. It uses non-transformer-based 2D feature extraction as well as 3D box prediction. 
It only uses multi-head attention to refine object queries by incorporating object interactions, similar to DETR \cite{carion2020end}.
PETR \cite{liu2022petr} first extracts features from multi-view images using a 2D backbone network (ResNet). 
It then uses the camera frustum space to generate a 3D mesh grid and coordinates in 3D space. 2D image features and 3D coordinates are then fused using an MLP-based encoder to generate 3D position-aware features.  
A transformer decoder then updates object queries based on their interaction with the 3D position-aware features.

\begin{table*}[t]
\renewcommand{\arraystretch}{0.5}
    \centering
    \tiny
    \caption{Overview of 3D segmentation methods using the transformer architecture. We divide 3D segmentation methods into three categories: (1) methods that perform 3D semantic segmentation on complete scenes (rather than single object part segmentation), (2) panoptic segmentation, and (3) medical imaging segmentation. 3D medical images are represented by a dense regular grid, in contrast to data collected by LiDARs and RGB-D sensors which is sparse.}
%        \begin{tabular}{|p{0.1\textwidth}|p{0.1\textwidth}
%    |p{0.13\textwidth}|p{0.05\textwidth}
%    |p{0.08\textwidth}|p{0.38\textwidth}|}
            \begin{tabular}{p{0.12\textwidth}p{0.09\textwidth}
    p{0.14\textwidth}p{0.05\textwidth}
    p{0.06\textwidth}p{0.40\textwidth}}
    \toprule     \belowrulesepcolor{gray!30!} 
\rowcolor{gray!30!} Method & Input & Scalability Element & Architecture & Context & Highlight \\ \aboverulesepcolor{gray!30!} \midrule \belowrulesepcolor{gray!30!}
\rowcolor{gray!30!} \multicolumn{6}{c}{\textbf{Complete scenes   segmentation:}} \\ \aboverulesepcolor{gray!30!} \midrule
Fast Point Transformer   \cite{park2021fast} & voxels & centroid aware voxelization & pure & local & speed-up local self-attention networks with voxel hashing   architecture and centroid-aware voxelization and devoxelization \\  \midrule
Stratified Transformer   \cite{lai2022stratified} & points & local aggregation with KPConv & hybrid & local & transformer-based hierarchical structure with a stratified   strategy for keys sampling \\ \midrule
Segment-Fusion   \cite{thyagharajan2022segment} & points & group points to 3D segments & hybrid & global & graph segmentation to segment features, segment fusion using   attention encoder block \\ \midrule
P4Transformer   \cite{fan2021point} & point cloud video & temporal radius \& stride, spatial radius \& subsampling & hybrid & global & extracts features of sampled local spatio-temporal using 4D   convolution, then use feature vector as input to transformer \\ \midrule
Wei \etal   \cite{wei2022spatial} & point cloud video & FPS then set abstraction & hybrid & global & set abstraction \& convolution layers to obtain patch features which are input to a transformer \\ \midrule \belowrulesepcolor{gray!30!}
\rowcolor{gray!30!}\multicolumn{6}{c}{\textbf{Panoptic:}} \\ \aboverulesepcolor{gray!30!} \midrule
Xu \etal  \cite{xu2022sparse} & points, voxels & voxelization & hybrid & global & sparse cross-scale attention network that aggregates sparse features at multiple scales and global voxel-encoded attention \\ \midrule \belowrulesepcolor{gray!30!}
\rowcolor{gray!30!} \multicolumn{6}{c}{\textbf{Medical imaging   segmentation}} \\ \aboverulesepcolor{gray!30!} \midrule
UNETR   \cite{hatamizadeh2022unetr} & 3D MRI images & patches & hybrid & global & patch embedding as input to the transformer, skip connection   between encoder and decoder \\ \midrule
D-Former \cite{wu2022d} & 3D medical images & patches & hybrid & local / global & Dilated Transformer that applies self-attention alternately in   local and global scopes \\ \midrule
CoTr \cite{xie2021cotr} & 3D medical images & from CNN encoder & hybrid & global & CNN encoder, multi-scale deformable self-attention, CNN   decoder \\ \midrule
T-AutoML \cite{yang2021t} & 3D CT scans & encodings of architectures, augmentation, hyperparameters & hybrid & - & Transformer to adapt to dynamic length to search for the best network architecture \\ \midrule
% T-AutoML \cite{yang2021t} & 3D CT & encodings of architectures, augmentation, hyperparameters & hybrid & global but not applicable & Transformer to adapt to dynamic length to search for the best network architecture \\ \midrule
Transfuse   \cite{zhang2021transfuse} & 3D medical images & patches & hybrid & global & fusion between CNN features and transformer features at   multiple scales \\ \midrule
Karimi \etal   \cite{karimi2021convolution} & 3D medical images & patches & pure & global & 3D patches as input to pure transformer \\ \midrule
SpecTr   \cite{yun2021spectr} & 3D medical images & from CNN encoder & hybrid & global & depth-wise convolution, spectral normalization, and   transformers as encoder \\ \midrule
TransBTS   \cite{wang2021transbts} & 3D MRI images & from CNN encoder & hybrid & global & transformer at 3D UNet bottleneck \\ \midrule
Segtran   \cite{li2021medical} & 3D medical images & FPN, from CNN encoder & hybrid & global & CNN with FPN as input to Squeeze-and-Expansion Transformer \\ \midrule
nnFormer   \cite{zhou2021nnformer} & 3D medical images & from CNN embedding & hybrid & local / global & local and global self-attention on CNN embedding with skip   self-attention \\ \midrule
BiTr-UNet   \cite{jia2021bitr} & 3D medical images & from CNN encoder & hybrid & global & Transformer at 3D UNet bottleneck, CBAM for 3D CNN \\ \midrule
AFTer-UNet   \cite{yan2022after} & 3D volume & from CNN encoder & hybrid & ~global & axial fusion transformer to fuse inter-slice and intra-slice   information \\ \midrule
Peiris \etal   \cite{peiris2021volumetric} & 3D volume & patches & pure & local / global & encoder with local/global attention, decoder with parallel window-based self/cross attention \\ \midrule
Swin UNETR   \cite{hatamizadeh2022swin} & 3D medical images & patches & hybrid & global & 1D sequence embedding as input to a Swin Transformer   \cite{liu2021swin} \\ \bottomrule
\end{tabular}
\label{tab:segmentation}
\vspacefigtext
\end{table*}

\vspacesection
\subsection{3D Segmentation}
3D segmentation aims at segmenting the 3D data constituting elements based on given semantic categories. 
It needs to overcome various challenges, such as class imbalance, size variation, and shape variation. 
In this section, we categorize methods based on the input data domain. 
We first review methods that take as input a 3D representation of a single object and then segment its parts. %
Next, we survey approaches that segment complete scenes with multiple objects.
We then present methods that provide point cloud segmentation from videos. 
In addition, we review methods that segment 3D medical images. 
An overview of the transformer design for these methods is shown in Table \ref{tab:segmentation}.

%\subsubsection*{Object Part Segmentation}
\vspace{1mm}
\noindent \textit{Object Part Segmentation.}
% semantic and instance segmentation
To perform 3D semantic segmentation on a given point, a label for every point is required. 
Numerous methods that learn point-wise features for point cloud classification can be employed for semantic segmentation.
%by using the per-point features instead of pooling them. 
This procedure is common in part-segmentation methods with the relatively smaller point clouds. 
Numerous transformer models have been developed for point cloud classification and semantic segmentation including: Attentional shapecontextnet \cite{xie2018attentional}, Yang \etal \cite{yang2019modeling}, Point2Sequence \cite{liu2019point2sequence}, Point Transformer \cite{zhao2021point}, Point Cloud Transformer (PCT) \cite{guo2021pct}, Point-Voxel Transformer (PVT) \cite{zhang2021pvt}, Adaptive wavelet Transformer \cite{huang2021adaptive}, and Dual Transformer \cite{han2021dual}. % please refer to classification section for more details

% \begin{figure*}[t]
% \renewcommand{\arraystretch}{0.5}
% \centering
% \begin{tabular}{c}
% \includegraphics[width=0.9\linewidth]{images/stratified.png} \\
% \end{tabular}
% \caption{Stratified Transformer for 3D point cloud segmentation: A transformer implementation that is applied locally with two properties to increase the receptive field: (1) A hierarchical structure is used to extract multi-level features, and (2) a stratified self-attention (SSA) block samples distant points as keys in a sparse way. Figure from \cite{lai2022stratified} (\copyright 2022 IEEE).}
% \label{stratified}
% \vspacefigtext
% \end{figure*}

\begin{figure}[t]
\renewcommand{\arraystretch}{0.5}
\centering
\begin{tabular}{c}
\includegraphics[width=0.75\linewidth]{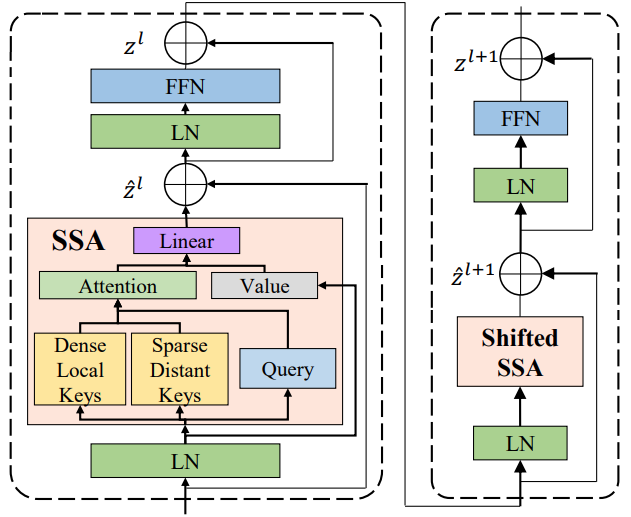} \\
\end{tabular}
\caption{Stratified Transformer for 3D point cloud segmentation: A transformer implementation applied locally with two properties to increase the receptive field: (1) A hierarchical structure is used to extract multi-level features, and (2) a stratified self-attention (SSA) block samples distant points as keys in a sparse way. Figure from \cite{lai2022stratified}.}
\label{stratified}
\vspacefigtext
\end{figure}

%\subsubsection*{Complete Scenes Segmentation}
\vspace{1mm}
\noindent \textit{Complete Scenes Segmentation.}
Fast Point Transformer \cite{park2021fast} proposes to speed up local self-attention networks as each such module 
usually requires finding the k-nearest points, which is computationally expensive. 
The proposed self-attention module learns on a point cloud with voxel hashing architecture. 
The voxel hashing allows fast neighborhood selection and is coupled with a centroid-aware voxelization and devoxelization to preserve the embedding of continuous coordinates.
On the other hand, Stratified Transformer	\cite{lai2022stratified} first aggregates local structure information using a point embedding module \cite{thomas2019kpconv}. 
It then uses a transformer-based hierarchical structure with several downsampling layers to obtain multi-level features, which are then upsampled layer by layer similar to U-Net \cite{ronneberger2015u}. 
The proposed transformer uses a stratified strategy for key sampling to increase the receptive field and aggregate long-range contexts (see Fig. \ref{stratified}).
Segment-Fusion	\cite{thyagharajan2022segment} employs graph segmentation methods to group points and their respective features into segments with segment-wise features. 
These features are then fused using a stack of attention encoder blocks, in which the attention matrix is also multiplied with the adjacency matrix to account for the connections between segments. 
The output of the attention-based is then grouped into object instances using the connected component algorithm.

% panoptic:
Transformers have also been applied to panoptic segmentation on point clouds. 
Xu \etal \cite{xu2022sparse} first generate point-wise features and sparse voxel features for a given point cloud. 
Voxel features are then aggregated using a cross-scale attention module, which allows capturing long-range relationship of object context and increases regression accuracy for the over-segmented large objects.

%\subsubsection*{Point Cloud Video Segmentation}
\vspace{1mm}
\noindent \textit{Point Cloud Video Segmentation.}
P4Transformer \cite{fan2021point} applies a transformer to point cloud videos for 3D action recognition and 4D semantic segmentation. 
It first samples and constructs local spatio-temporal areas, and uses a 4D convolution to encode them into a feature vector that can be processed by a transformer. 
In \cite{wei2022spatial}, Wei \etal~first extract features using set abstraction layers from \cite{qi2017pointnet++} and use a resolution embedding module to retain geometric information in the extracted features. 
It then applies a convolution on the features of neighboring frames to group them into patches. The patches are used as input to a spatio-temporal transformer to capture context information for the tasks of 3D action recognition and 4D semantic segmentation.

%\subsubsection*{3D Medical Images Segmentation}
\vspace{1mm}
\noindent \textit{3D Medical Images Segmentation.}
UNETR \cite{hatamizadeh2022unetr} divides the input 3D volume into a sequence of uniform non-overlapping patches and projects them into an embedding space using a linear layer.
A transformer is then applied to learn sequence representations of the input volume (encoder) and  capture the global multi-scale information. %The transformer encoder is directly connected to a decoder via skip connections at different resolutions to compute the final semantic segmentation output.
In \cite{xie2021cotr}, CoTr uses a CNN-encoder to extract multi-scale feature maps from an input 3D medical image. 
The feature maps are embedded with positional encoding and processed using a deformable self-attention transformer. % self-attention-based
The features are then upsampled to the original resolution using a CNN decoder.

In \cite{yang2021t}, T-AutoML introduces an automated search algorithm for finding the best network architecture, hyperparameters, and augmentation methods for lesion segmentation in 3D CT images.
It uses the transformer model for its ability to operate on varying embedding lengths.
On the other hand, D-Former \cite{wu2022d} proposes local and global attention-based modules to increase the scopes of information interactions without increasing the number of patches. 
A Dilated Transformer applies self-attention for pair-wise patch relations captured in the local and global scopes. 
It also applies dynamic position encoding to embed relative and absolute position information.

Recently, Transfuse \cite{zhang2021transfuse} presents a fusion module to fuse information from two branches: 
a CNN branch which encodes features from local to global and the other one is a transformer branch which starts with global self-attention and then recovers local information.
In \cite{karimi2021convolution} Karimi \etal~first divide the input 3D image block into 3D patches. 
A one-dimensional embedding is then computed for each patch and passed through an attention-based encoder to predict segmentation of the center patch.
SpecTr \cite{yun2021spectr} takes as input a sequence of spectral images and then alternately processes them with depth-wise convolution, spectral normalization, and transformers with sparsity constraint in the encoder.
It then uses a decoder with skip connections similar to 3D U-Net \cite{cciccek20163d}.
On the other hand, TransBTS \cite{wang2021transbts} uses a 3D CNN to generate feature maps that capture spatial and depth information, and then utilizes a transformer encoder to model the long-distance global contextual dependency. 
The transformer output is then alternately upsampled, stacked, and convoluted to produce the segmentation labels.
Segtran \cite{li2021medical} uses CNN layers to extract features, which are used as input to Squeeze-and-Expansion transformer layers to learn the global context. 
A Feature Pyramid Network  is applied before the transformer to increase the spatial resolution, and after the transformer to upsample to the original resolution.

%HERE
Zhou \etal~combine convolution and self-attention operations by applying local and global volume-based self-attention operations in nnFormer \cite{zhou2021nnformer}.
It also proposes to use skip attention which is analogous to skip connections in UNet-like architectures.
BiTr-UNet \cite{jia2021bitr} applies a transformer block at the bottleneck of a 3D UNet architecture i.e., after the 3D CNN encoder and before the upsampling layers. 
It integrates CBAM \cite{woo2018cbam} into the convolution layers by expanding it for 3D CNN.
AFTer-UNet \cite{yan2022after} encodes neighboring slice groups using a CNN encoder and then applies an axial fusion transformer. The axial fusion transformer fuses inter-slice and intra-slice information which is then passed into a CNN decoder for segmentation.
Peiris \etal \cite{peiris2021volumetric} propose an encoder block design with local and global self-attention layers. 
It uses a decoder with window-based self and cross attention, where both attention mechanisms use one shared query projection.
In addition, it proposes a convex combination approach in the decoder along with Fourier positional encoding.
Swin UNETR \cite{hatamizadeh2022swin} projects multi-modal input data into a 1D sequence of embedding and uses it as input to an encoder composed of hierarchical Swin Transformer \cite{liu2021swin}. 
The Swin Transformer uses shifted windows to compute self-attention at multiple resolutions and has skip connections to a FCNN decoder.

\begin{table}[t]
\renewcommand{\arraystretch}{0.5}
    \centering
    \tiny
    %MH2: check this caption
    \caption{State-of-the-art 3D point cloud completion methods using transformers. These methods use a variety of input representations, employ a pure or hybrid  architecture, and apply the transformer locally or globally.}
          \begin{tabular}{lllp{0.035\textwidth}p{0.19\textwidth}}
    \toprule     \belowrulesepcolor{gray!30!}
\rowcolor{gray!30!} Method & Input & Architecture &  Context & Highlight \\  \aboverulesepcolor{gray!30!} \midrule
    PoinTr \cite{Yu2021PoinTr}  & points  & hybrid  & global/ local &  a set-to-set translation  and geometry-aware transformer for point cloud completion\\ \midrule
    Wang \etal \cite{Wang2022localdisp}  & points & hybrid & global/ local  & neighboring pooling integrated with transformer encoder-decoder\\ \midrule 
    PointAttN \cite{Wang2022PointAttN}  & points  & hybrid  & global/ local  & a geometric details perception module and a self-feature augment module\\ \midrule     
    SnowflakeNet \cite{Xiang2021SnowflakeNet}  & points  & hybrid & local  & a skip-transformer module to capture shape context\\ \midrule  
    Su \etal \cite{Su2022SFMFN}  & points  & hybrid & global/ local  & attention for feature aggregation and upsampling\\ \midrule      
    PCTMA-Net \cite{Lin2021PCTMANet}  & points  & hybrid & global  & a transformer encoder to learn semantic affinity information\\ \midrule   
    AutoSDF \cite{Mittal2022AutoSDF}  & voxel & hybrid & global  & transformer-based autoregressive modeling in  low-dimension latent space\\ \midrule
    ShapeFomer \cite{Yan2022ShapeFormer}  & voxel & hybrid & global  & a transformer to predict a distribution of object completions\\ \midrule   
    PDR \cite{Lyu2022DDPM}  & point & hybrid & local  & aggregating local features by the attention operation\\ \midrule
    MFM-Net \cite{Cao2021MFM-Net}  & point & hybrid & global  & a self-attention layer for global refinement\\ \bottomrule 
\end{tabular}
\label{tab:comp}
\vspacefigtext
\end{table}

\begin{figure}[t]
\renewcommand{\arraystretch}{0.8}
% \label{table_example}
\centering
\begin{tabular}{c}
\includegraphics[width=\linewidth]{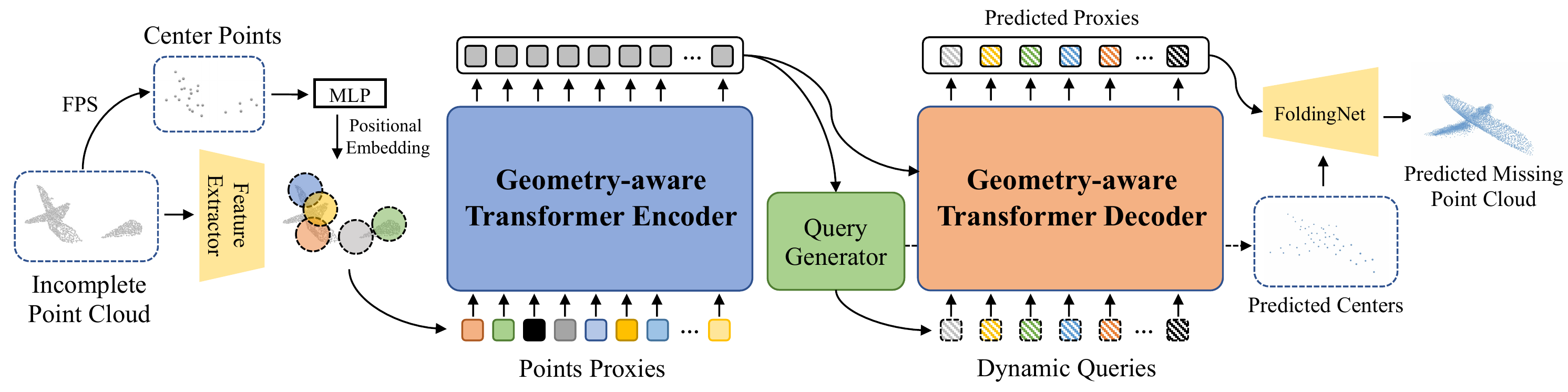} \\
\end{tabular}
\caption{PoinTr \cite{Yu2021PoinTr}: Geometry-aware transformer for point cloud completion. PoinTr first extracts the deep features of center points, second adopts the geometry-aware transformer encoder-decoder to predict the proxies of missing points, and third employs FoldingNet to complete the point cloud. The figure is from \cite{Yu2021PoinTr} (\copyright 2021 IEEE).}
\label{fig:pointr}
\vspacefigtext
\end{figure}

%MH: is this paragraph written by someone else? The style (e.g., tense of verb) is different from other paragraphs. I change all past tense to current tense (similar to all the other paragraphs). This paragraph sometimes describes a method using one sentence. While it is fine in some cases, you still need to give sufficient detail. More importantly, it is important to give insight (not a list of papers with one-sentence summary). 
% add more details
\vspacesection
\subsection{3D Point Cloud Completion}
Point cloud completion aims to generate a complete point cloud from a partial point cloud of an object. 
Table \ref{tab:comp} gives a brief summary of some 3D point cloud completion methods using the transformer structure. 
%
%Some methods adopt transformer encoder-decoder structure. 
%
Fig. \ref{fig:pointr} shows the architecture of PoinTr \cite{Yu2021PoinTr}. 
%
%MH: need to give a bit more detail. It reads like you simply give one entry
% add more
PoinTr \cite{Yu2021PoinTr} treats point cloud completion as a set-to-set translation task and adopts a geometry-aware transformer to predict the missing point cloud. 
Compared to the vanilla transformer, the geometry-aware transformer contains two branches, where one branch adopts self-attention to extract semantic features and another branch adopts kNN model to extract geometric features. 
The output features of two branches are fused to generate the output feature of geometry-aware transformer.
%
%MH: need to give a bit more detail
%add more about down-sampling
Wang \etal \cite{Wang2022localdisp} integrate the proposed down-sampling and up-sampling operations into the transformer encoder-decoder structure \cite{Yu2021PoinTr} to perform point cloud completion. 
Instead of max pooling, the down-sampling operation adopts neighbor pooling to select features with the highest activations.
In \cite{Wang2022PointAttN}, Wang \etal~introduce a geometric details perception module and a self-feature augment module to capture both local and global information and avoid local k-nearest neighbor operation. 

We note that some methods only use a transformer encoder or decoder for 3D point cloud completion. 
Xiang \etal \cite{Xiang2021SnowflakeNet} propose a skip-transformer module in the decoder to capture context for sparse point deconvolution.  The skip-transformer aims to use an attention mechanism to learn spatial context from the previous decoding layer during point generation stage.
Recently, Su \etal \cite{Su2022SFMFN} employ a transformer encoder to extract context information from coarse point cloud generation.  After that, a point cloud upsampling is used to generate a fine point cloud.
In \cite{Lin2021PCTMANet}, Lin \etal~utilize a transformer encoder to learn semantic affinity information, which is followed by a morphing-atlas point generation decoder. Some methods consider reducing the computational costs when applying a transformer for point cloud completion. 
Mittal \etal \cite{Mittal2022AutoSDF} propose to first map the high-dimension 3D shape to low-dimension latent space and second perform transformer-based autoregressive modeling. 
On the other hand, Yan \etal \cite{Yan2022ShapeFormer} use a novel vector quantized deep implicit functions to sparsely encode 3D shape and employ a transformer module to predict conditional distribution of location and content. 

Some methods adopt the attention module in the transformer to help feature extraction.
For example, Lyu \etal \cite{Lyu2022DDPM} adopt the attention operation, instead of the pooling operation, to aggregate local features. 
%
%MH: need a bit more detail. 
%added
Cao \etal \cite{Cao2021MFM-Net} add a self-attention layer in global refinement module that is used to enhance the details and avoid distortion.

\begin{figure}[t]
\renewcommand{\arraystretch}{0.8}
% \label{table_example}
\centering
\begin{tabular}{c}
\includegraphics[width=0.95\linewidth]{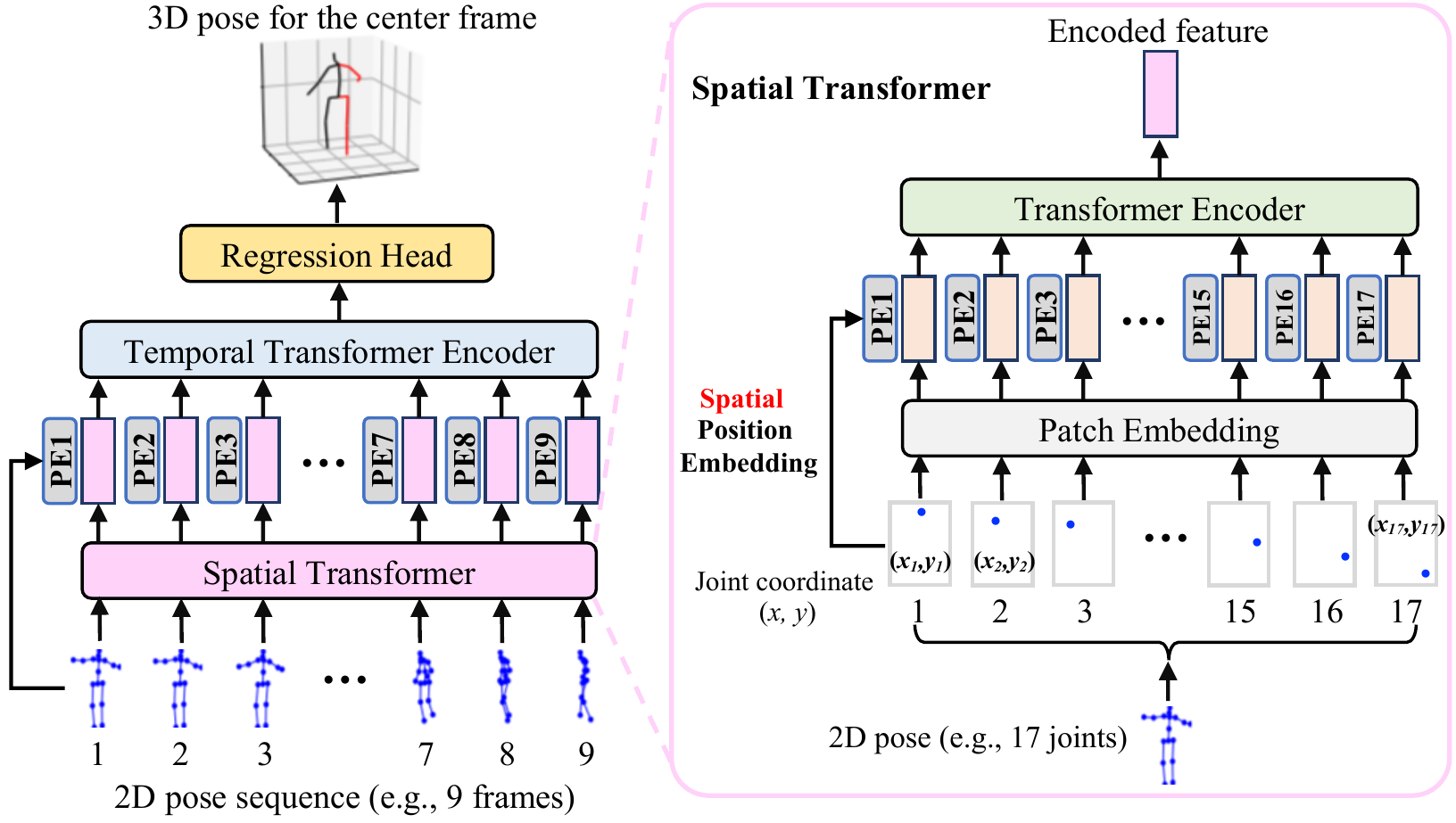} \\
\end{tabular}
\caption{PoseFormer \cite{Zheng2021PoseFormer}:  Transformer-based approach for 3D
human pose estimation in videos. PoseFormer takes the 2D pose sequence of multiple frames, generated by an off-the-shelf 2D pose detector, as the input. After that, PoseFormer employs a spatial-temporal transformer to exploit the local and global information of different poses. Finally, a regression head is used to predict the 3D pose of the center frames.  The figure is from \cite{Zheng2021PoseFormer} (\copyright 2021 IEEE).}
\label{fig:poseformer}
\vspacefigtext
\end{figure}

\begin{table}[t]
\renewcommand{\arraystretch}{0.5}
    \centering
    \tiny
    \caption{Summary of 3D pose estimation related approaches using the transformer. Some methods focus on videos or multi-frames, some methods focus on multi-view frames, some methods take single depth or RGB image as input, and some methods are for 6D pose estimation.}
%    \begin{tabular}{|p{0.15\textwidth}|p{0.07\textwidth}
%    |p{0.09\textwidth}|p{0.075\textwidth}|p{0.5\textwidth}|}
        \begin{tabular}{p{0.11\textwidth}p{0.03\textwidth}
    p{0.03\textwidth}p{0.025\textwidth}p{0.18\textwidth}}
    \toprule     \belowrulesepcolor{gray!30!}
\rowcolor{gray!30!} Method & Input & Arch. & Context & Highlight \\ \aboverulesepcolor{gray!30!} \midrule
 \multicolumn{5}{l}{Videos or multi-frames}   \\ \midrule
PoseFormer \cite{Zheng2021PoseFormer}  & sequence & pure & global/ local &  a spatial-temporal transformer encoder module\\ \midrule
Crossformer \cite{Hassanin2022CrossFormer}  & sequence & pure & global/ local &  a cross-joint interaction module and a cross-frame interaction module\\ \midrule
STE \cite{Li2022STE}  & sequence & pure & global/ local &  vanilla and strided transformers for global and local information aggregation\\ \midrule  
P-STMO \cite{Shan2022P-STMO}  & sequence & pure & global/ local &  self-supervised pre-training technique with transformer  \\ \midrule 
MixSTE \cite{Zhang2022MixSTE}  & sequence & pure & global/ local &  spatio-temporal attention in an alternating style \\ \midrule  
MHFormer \cite{Li2022MHFormer}  & sequence & pure & global/ local &  spatio-temporal representations of
multiple pose hypotheses \\ \midrule   
GraFormer \cite{Zhao2022GraFormer}  & sequence & hybrid & global/ local & a novel
model by combining graph convolution and transformer \\ \midrule   
\midrule
 \multicolumn{5}{l}{Multi-view frames}   \\ \midrule
Epipolar \cite{He2022Epipolar}  & multi-view & hybrid & local &  enhance the feature of source view from and source view\\ \midrule  
RayTran \cite{Tyszkiewicz2022RayTran}  & multi-view & hybrid & local &  a ray-traced transformer to progressively exchange information\\
\midrule \midrule
 \multicolumn{5}{l}{Singe depth or RGB image}   \\ \midrule
Hand-transformer \cite{Huang2020Hand-Transformer}  & point & hybrid & global &  a non-autoregressive hand transformer to model the relationships of input points\\\midrule
Cheng \etal \cite{Cheng2022EVVS}  & point & hybrid & global &  an attention based confidence network for multi-view feature fusion\\\midrule  
METRO \cite{Lin2021METRO}  & image & hybrid & global &  a progressive dimensionality reduction transformer to predict 3D coordinates\\\midrule  
HOT-Net \cite{Huang2020HOT-Net}  & image & hybrid & global &  a hand-object transformer based on non-autoregressive transformer\\\midrule  
HandsFormer \cite{Hampali2021HandsFormer}  & image & hybrid & global &   self-attention between  the keypoints to associate them\\\midrule  
PI-Net \cite{Guo2021PI-Net}  & image & hybrid & global & a self-attention module to improve the embeddings of each person\\\midrule   
Ugrinovic \etal \cite{Ugrinovic2022PIRN}  & image & hybrid & global &  encode global information of multiple persons with set transformer\\\midrule  
\midrule 
 \multicolumn{5}{l}{6D pose estimation}   \\ \midrule
6D-ViT \cite{Zou20216D-ViT}  & image & pure & global &  a two-branch transformer encoder-decoder structure\\ \midrule
Dang \etal \cite{Dang2022DCP}  & image & hybrid & global &  a DCP-based architecture with transformer for feature extraction\\ \midrule
Goodwin \etal \cite{Goodwin2022ZSCL}  & image & pure & global &  zero-shot category-level pose estimation\\ \bottomrule
\end{tabular}
\label{tab:pose}
\vspacefigtext
\end{table}

%MH: the verbs of this section are also in the past tense.
%I can see that sometimes one author use \etal and another one uses \textit{et al.}. While the results are the same, it is best to use one consistent style (in case we need to revise it).
% change to \etal
\vspacesection
\subsection{3D Pose Estimation}
3D pose estimation aims to estimate 3D joint locations of objects from images or videos. 
Table \ref{tab:pose} summarizes the related 3D pose estimation methods using the transformer structure. 
%
%Some researchers explored spatio-temporal information fusion for video based pose estimation.
A few methods that fuse spatio-temporal information for video based pose estimation have been developed. 
%video human pose
Zheng \etal \cite{Zheng2021PoseFormer} propose a spatial-temporal transformer encoder module, PoseFormer, to model local relations within frame and global relations across frames.
Fig. \ref{fig:poseformer} shows the architecture of PoseFormer. 
Hassanin \etal
\cite{Hassanin2022CrossFormer} show that PoseFormer has the issue of poor locality and propose two novel cross interaction modules to integrate locality and inter interaction.
Specifically, a cross-joint interaction module is used to encode local part information within frame, and a cross-frame interaction module is to encode information of the joints across frames.
On the other hand, 
Li \etal \cite{Li2022STE} adopt a vanilla transformer to exploit long-range information  and design a strided transformer to progressively aggregate long-range information of different frames into a single 3D representation.
In \cite{Shan2022P-STMO}, Shan \etal~introduce a self-supervised pre-training method with transformer for 3D human pose estimation.
%
%MH: need some more detail (not much insight)
Li \etal \cite{Li2022MHFormer} propose to learn spatio-temporal representations of multiple pose hypotheses with the proposed multi-hypothesis transformer. In multi-hypothesis transformer, there are three modules: multi-hypothesis generation (MHG), self-hypothesis refinement (SHR), and cross-module interaction (CHI). MHG aims to explore spatial information within frame, and SHR and CHI aim to explore temporal information accross frames. 
Zhao \etal \cite{Zhao2022GraFormer} develop a graph-oriented transformer to model the relation between different joints, which integrates graph convolution and attention together. 
%
%
%MH: This sentence does not follow with the previous sentences,. It is also unclear, central frame?
% change
These methods above employ transformer to predict single 3D pose estimation of the central frame in the video (called seq2frame). 
Different from these approaches,  Zhang \etal \cite{Zhang2022MixSTE} present MixSTE for 3D pose sequence estimation that peforms 3D pose estimation for all frames in the video (called seq2seq).
MixSTE takes each 2D joint as a token and performs spatio-temporal attention in an alternating style.

%MH: a better way for segway is to describe the main them of the methods described in this paragraph. 
% You have two consecutive paragraphs starting with "some researchers..." boring and mechanical...
%multi-view pose estimation
%Some researchers explored multi-view pose estimation.
Instead of replying on single views, a few methods use multiple views for 3D pose estimation. 
He \etal \cite{He2022Epipolar} propose an epipolar transformer \cite{He2020Epipolar} to enhance the feature of reference view with the feature of the corresponding point in source view.
Tyszkiewicz \etal \cite{Tyszkiewicz2022RayTran} exploit features of multiple frames and empty volumetric feature as inputs, and employ a ray-traced transformer to progressively exchange information for 3D representation.

A few approaches estimate 3D pose from depth or RGB images. 
%
%MH: what is non-autoregressive? revise it
%depth feature
Huang \etal \cite{Huang2020Hand-Transformer} introduce a non-autoregressive hand transformer  (NARHT) to avoid sequential inference in transformer and achieve a fast inference speed. Inspired by non-autoregressive transformer \cite{Gu2018NAT}, NARHT uses a structured-reference extractor to predict reference pose and model the relationships of input points and reference pose fin parallel.
Cheng \etal \cite{Cheng2022EVVS} propose an attention based confidence network to predict the confidence of each virtual view and select the important views for depth-based hand  pose estimation.
In \cite{Lin2021METRO}, Lin \etal~present a progressive dimensionality reduction transformer to predict 3D coordinates of each joint for RGB-based human pose and mesh reconstruction. These methods above focus on single object pose estimation. 
Huang \etal \cite{Huang2020HOT-Net} develop a hand-object transformer network to leverage the joint correlations between hand and object for  hand-object
pose estimation.
In addition, some multi-object pose estimation methods are proposed.
Hampali \etal \cite{Hampali2021HandsFormer} first extract a set of keypoints and second took their appearance and spatial encodings as the input to transformer for 3D hand and object pose estimation. 
In \cite{Guo2021PI-Net}, Guo \etal~employ a self-attention module to improve the embeddings of each person by integrating the embeddings of other persons.
Recently,  Ugrinovic \etal \cite{Ugrinovic2022PIRN} exploit the set transformer \cite{Lee2019Set} to encode global information of multiple persons for improved multi-person pose estimation.

% 6D pose estimation
%In addition to 3D pose estimation, some researchers employed transformer for 6D pose estimation. 
%MH: what is 6D pose? explain it.
Aside from 3D pose estimation, a few models have been developed to describe 6D pose information. 
Zou and Huang \cite{Zou20216D-ViT} design a two-branch transformer encoder-decoder structure to respectively extract features from image and point cloud, and perform 6D pose estimation based on the aggregated features of two branches.
%
%MH: what is DCP?
Dang \etal \cite{Dang2022DCP} employ a learn-based point cloud registration for 6D pose estimation, where a transformer is used to combine the features from two point inputs in point cloud registration architecture DCP \cite{wang2019deep}.
Goodwin \etal \cite{Goodwin2022ZSCL} develop a novel zero-shot category-level 6D pose estimation task and employ a self-supervised transformer for feature extraction.

%MH: Section 4.5 has similar issues as Section 4.2. The sentences are basically one or two liners to list each method without much insight. 
% add more details
\vspacesection
\subsection{Other Tasks}
%\subsubsection*{3D Object Tracking}
\vspace{1mm}
\noindent \textit{3D Object Tracking.}
3D Point Cloud object tracking aims to localize the object in 3D space given a template point cloud. 
Recently, a few 3D tracking methods based on transformers have been developed. 
Cui \etal \cite{Cui2021LTTR} use a transformer to exploit local and global information within a point cloud and across different point clouds for 3D tracking prediction.
Similarly, Zhou \etal \cite{Zhou2022PTTR} employ a self-attention module to  capture long-range dependencies and a cross-attention module for a coarse matching.

%\subsubsection*{3D Motion Prediction}
\vspace{1mm}
\noindent \textit{3D Motion Prediction.}
3D motion prediction aims to predict the future pose based on a history of past motion.
Mao \etal \cite{Mao2020HisRepItself} propose a motion attention module to  capture the motion relation of the  long-term history of motion. 
Similar to 3D pose estimation, some approaches \cite{Cai2020dct,Aksan2021ST,Medjaouri2022HR-STAN} use transformers to capture spatial and temporal long-range dependency between joints. 
%
%MH: non-autoregressive transformer?
% add the reference
Gonzalez \etal~take pose sequences \cite{Gonzalez2021POTR} as the input and use non-autoregressive transformer \cite{Gu2018NAT} for motion prediction. 
Zheng \etal \cite{Zheng2022GIMO} propose an ego-centric motion prediction dataset and develop a cross-modal transformer module for this task.
 
%  Cai \textit{et al.} proposed a transformer-based architecture to capture the relation between different joints and different motions. Aksan \textit{et al.} 

%\subsubsection*{3D Reconstruction}
\vspace{1mm}
\noindent \textit{3D Reconstruction.} 
Wang \etal \cite{Wang2021MV3D} focus on multi-view 3D object reconstruction and proposed a 3D volume transformer for feature extraction and fusion.  
%
%MH: no sufficient detail
% add
Zanfir \etal \cite{Zanfir2021THUNDR} reconstruct human 3D shape from a monocular image with the THUNDR model. 
THUNDR predicts and regularizes an intermediate 3d  marker representation with CNN feature extraction and transformer refinement.  
%
%MH: no sufficient detail
Mahmud and Frahm \cite{Mahmud2022VPFusion} develop single-view and multi-view object reconstruction methods, termed as VPFusion, with transformer-based feature fusion. VPFusion adopts transformer to perform cross-view  feature fusion.

%\subsubsection*{Point Cloud Registration}
\vspace{1mm} 
\noindent \textit{Point Cloud Registration.}
DCP \cite{wang2019deep} uses self-attention and conditional attention to approximate combinatorial matching between two points clouds. 
In  \cite{fu2021robust}, Fu \etal~introduce a point cloud registration technique based on graph matching. % deep graph matching
It utilizes the transformer to generate edges for the graph construction. 
REGTR \cite{yew2022regtr} employs the transformer for point cloud registration by predicting the probability of each point to lie in the overlapping region of two scans.

\begin{table}[t]
\tiny
\renewcommand{\arraystretch}{0.5}
    \centering
    \caption{Common datasets for 3D vision tasks. These datasets cover a range of 3D applications: classification, segmentation, detection, completion, and pose estimation. They are collected using different sensors for indoor scenes, outdoor scenes, and objects.}
%    \begin{tabular}{|l|l|l|l|l|}
        \begin{tabular}{lllll}
    \toprule     \belowrulesepcolor{gray!30!}
        \rowcolor{gray!30!} Dataset & Task & Size & Cls & Note  \\ \aboverulesepcolor{gray!30!} \midrule
        ModelNet40 \cite{wu20153d}  & classification & 12,311 models & 40 & CAD models  \\ \midrule
        ShapeNet \cite{chang2015shapenet}  & segmentation & 16,880 models & 16 & 50 different parts\\ \midrule
        S3DIS \cite{armeni2017joint}  & segmentation & 271 rooms  & 13 &   six areas  \\ \midrule
        ScanObjectNN \cite{scanobjectnn}  & classification & 2902 point clouds & 15 & with background/occlusions \\ \midrule
        SUN RGB-D \cite{song2015sun}  & detection & 10,335 frames & 37 & Over 64,000 3D BB  \\ \midrule
        ScanNet \cite{dai2017scannet}  & segm., det. & 1513 scenes & 40 & 100 hidden test scenes  \\ \midrule
        KITTI \cite{Geiger2012CVPR}  & detection & 14,999 frames & 3 & 80,256 labeled objects  \\ \midrule
        NuScenes \cite{nuscenes}  & detection & 1k scenes & 23 & 40K key frames, 8 attributes  \\ \midrule
        Completion3D \cite{Tchapmi2019com3d}  & completion  &  30,958 models &  8 & same size of 2048 × 3     \\ \midrule
        PCN \cite{Yuan2018pcn} & completion  &  30,974 models &  8 & less than 2048 points \\ \midrule
        Human3.6M \cite{Ionescu2014Human36m}  & pose estimation  & 3.6M video frames  &  17 & Indoor with 11 actors \\ \midrule
        MPI-INF-3DHP \cite{Mehta20173DHP}  &  pose estimation &  14 camera views  &  8 & indoor/outdoor with 8 actors \\ \bottomrule
    \end{tabular}
    \label{tab:dataset}
    \vspacefigtext
\end{table}

\vspacesection
\section{Benchmark Performance}\label{sec:benchmark}

To understand the effect of incorporating the transformer architecture into 3D vision pipelines, it is essential to compare against previous methods. 
In this section, we present quantitative comparisons of transformer-based architecture to state-of-the-art non-transformer methods on benchmark datasets. 
We start by providing details on the datasets and evaluation metrics, and then show the quantitative results for different tasks.

\begin{table}[t]
\renewcommand{\arraystretch}{0.5}
    \centering
    \tiny
    \caption{Shape classification results on ModelNet40 benchmark (P: points, N:normals, *: pre-trained).}
%        \begin{tabular}{|p{0.18\textwidth}|p{0.04\textwidth}|p{0.05\textwidth}|p{0.05\textwidth}}
        \begin{tabular}{p{0.15\textwidth}p{0.08\textwidth}p{0.08\textwidth}p{0.08\textwidth}}
    \toprule \belowrulesepcolor{gray!30!}   
\rowcolor{gray!30!} Method & Repr. & \# points & Acc (\%) \\ \aboverulesepcolor{gray!30!} \midrule \belowrulesepcolor{gray!30!}
\rowcolor{gray!30!} \multicolumn{4}{c}{\textbf{Non-Transformer methods:}} \\ \aboverulesepcolor{gray!30!} \midrule
PointNet   \cite{qi2017pointnet} & P & 1k & 89.2 \\ \midrule
PointNet++   \cite{qi2017pointnet++} & P + N & 5k & 91.9 \\ \midrule
PointCNN   \cite{li2018pointcnn} & P & 1k & 92.5 \\ \midrule
KPConv   \cite{thomas2019kpconv} & P & 6.8k & 92.9 \\ \midrule
DGCNN   \cite{wang2019dynamic} & P & 1k & 92.9 \\ \midrule
RS-CNN   \cite{liu2019relation} & P & 1k & 93.6 \\ \midrule \belowrulesepcolor{gray!30!}
\rowcolor{gray!30!} \multicolumn{4}{c}{\textbf{Transformer-based methods:}} \\ \aboverulesepcolor{gray!30!} \midrule
A-ShapeContextNet   \cite{xie2018attentional} & P & 1k & 90.0 \\ \midrule
Yang \etal \cite{yang2019modeling} & P & 1k & 91.7 \\ \midrule
3DETR \cite{misra2021end} & P &  & 91.9 \\ \midrule
Point Transformer   \cite{engel2021point} & P & 1k & 92.8 \\ \midrule
MLMSPT   \cite{han2021point} & P & 1k & 92.9 \\ \midrule
DTNet \cite{han2021dual} & P & 1k & 92.9 \\ \midrule
Wu \etal   \cite{wu2021centroid} & P & 1k & 93.2 \\ \midrule
LFT-Net \cite{gao2022lft} & P + N & 2k & 93.2 \\ \midrule
PCT \cite{guo2021pct} & P & 1k & 93.2 \\ \midrule
3DCTN \cite{lu20223dctn} & P + N & 1k & 93.3 \\ \midrule
3DMedPT \cite{yu20213d} & P & 1k & 93.4 \\ \midrule
3CROSSNet   \cite{han20223crossnet} & P & 1k & 93.5 \\ \midrule
PAT   \cite{cheng2021patchformer} & P + N & 1k & 93.6 \\ \midrule
Point Transformer   \cite{zhao2021point} & P + N & 1k & 93.7 \\ \midrule
Point-BERT   \cite{yu2021point} & P & 8k & 93.8* \\ \midrule
Point-MAE   \cite{pang2022masked} & P & 1k & 93.8* \\ \midrule
Adaptive Wavelet \cite{huang2021adaptive} & P & 1k & 93.9 \\ \midrule
CpT \cite{kaul2021cpt} & P & 1k & 93.9 \\ \midrule
Liu \etal   \cite{liu2022group} & P & 1k & 93.9 \\ \midrule
PVT \cite{zhang2021pvt} & P + N & 1k & 94.0 \\ \bottomrule
\end{tabular}
\label{tab:modelnet}
\vspacefigtext
\end{table}

\begin{table}[t]
\renewcommand{\arraystretch}{0.4}
    \centering
    \tiny
    \caption{Object classification results on ScanObjectNN dataset \cite{scanobjectnn}. We report classification accuracy on the three main variants: OBJ-BG, OBJ-ONLY, and PB-T50-RS }
        \begin{tabular}{p{0.15\textwidth}p{0.08\textwidth}p{0.08\textwidth}p{0.08\textwidth}}
    \toprule \belowrulesepcolor{gray!30!}   
\rowcolor{gray!30!} Method    & OBJ-BG & OBJ-ONLY & PB-T50-RS  \\
\aboverulesepcolor{gray!30!} \midrule \belowrulesepcolor{gray!30!}
\rowcolor{gray!30!} \multicolumn{4}{c}{\textbf{Non-Transformer methods:}} \\ \aboverulesepcolor{gray!30!} \midrule
PointNet++   \cite{qi2017pointnet++} & 82.3   & 84.3     & 77.9      \\ \midrule
DGCNN   \cite{wang2019dynamic}       & 82.8   & 86.2     & 78.1 \\ \midrule
\belowrulesepcolor{gray!30!}
\rowcolor{gray!30!} \multicolumn{4}{c}{\textbf{Transformer-based methods:}} \\ \aboverulesepcolor{gray!30!} \midrule
Point-BERT   \cite{yu2021point}      & 87.43  & 88.12    & 83.07     \\ \midrule
Pang \etal   \cite{pang2022masked}   & 90.02  & 88.29    & 85.18     \\ \bottomrule
\end{tabular}
\label{tab:scanobjectnn}
\vspacefigtext
\end{table}

\vspacesection
\subsection{Datasets}
Datasets play two important roles in the advancement of computer vision tasks.
First, annotations allow training deep learning models to solve challenging problems. 
Second, datasets provide a basis for quantitative comparisons to measure the effectiveness of a proposed method. 
Numerous datasets have been developed for 3D vision tasks. 
For classification of 3D objects, ModelNet40 \cite{wu20153d} is widely used, as well as ScanObjectNN \cite{scanobjectnn}. 
For 3D segmentation, ShapeNet \cite{chang2015shapenet} provides part segmentation annotation, and S3DIS \cite{armeni2017joint} and ScanNet \cite{dai2017scannet} provide indoor scenes segmentation. 
For indoor 3D object detection, SUN RGB-D \cite{song2015sun} provides oriented bounding boxes annotations for scenes from a single RGB-D frame, and ScanNet \cite{dai2017scannet} can also be utilized by transforming instance segmentation labels into axis aligned bounding box annotations. For 3D object detection in outdoor scenes, KITTI \cite{Geiger2012CVPR} and nuScenes \cite{nuscenes} are the most commonly used.
For 3D point cloud completion, Completion3D \cite{Tchapmi2019com3d} and PCN \cite{Yuan2018pcn} are two of the most widely used datasets. 
For 3D pose estimation, Human3.6M \cite{Ionescu2014Human36m} and MPI-INF-3DHP \cite{Mehta20173DHP} are among the most popular datasets.
The main attributes of these datasets are shown in Table \ref{tab:dataset}.

\begin{table}[t]
\renewcommand{\arraystretch}{0.5}
    \centering
    \tiny
    \caption{Object Part Segmentation on ShapeNet \cite{chang2015shapenet} and scene semantic segmentation results on S3DIS \cite{armeni2017joint}. We compare transformer-based methods to state-of-the-art non-transformer methods. For part segmentation, we show instance mean IoU, whereas for scene segmentation, we report mean accuracy (mAcc) and mean IoU.}
     \begin{tabular}{p{0.15\textwidth}p{0.08\textwidth}
    p{0.08\textwidth}p{0.08\textwidth}}
    \toprule \belowrulesepcolor{gray!30!}
\rowcolor{gray!30!} ~ & ShapeNet & \multicolumn{2}{c}{S3DIS} \\ \aboverulesepcolor{gray!30!} \midrule \belowrulesepcolor{gray!30!}
\rowcolor{gray!30!} Method & ins. mIoU & mAcc & mIoU \\ \aboverulesepcolor{gray!30!} \midrule \belowrulesepcolor{gray!30!}
\rowcolor{gray!30!} \multicolumn{4}{c}{\textbf{Non-Transformer methods:}} \\ \aboverulesepcolor{gray!30!} \midrule
ShapeNet   \cite{chang2015shapenet} & 81.4 & - & -  \\ \midrule
PointNet   \cite{qi2017pointnet} & 83.7 & 49.0 & 41.1 \\ \midrule
PointNet++   \cite{qi2017pointnet++} & 85.1 & - & 51.5 \\ \midrule
DGCNN   \cite{wang2019dynamic} & 85.2 & - & 56.1  \\ \midrule 
MinkowskiNet   \cite{choy20194d} & – & 71.7 & 65.4 \\ \midrule 
KPConv   \cite{thomas2019kpconv} & 86.4 & 72.8 & 67.1 \\ \midrule \belowrulesepcolor{gray!30!}
\rowcolor{gray!30!} \multicolumn{4}{c}{\textbf{Transformer-based methods:}} \\ \aboverulesepcolor{gray!30!} \midrule
Point Transformer  \cite{zhao2021point} & 86.6 & 76.5 & 70.4 \\ \midrule
Point Transformer   \cite{engel2021point} & 85.9 & - & - \\ \midrule
A-ShapeContext \cite{xie2018attentional} & 84.6 & - & 52.7 \\ \midrule
Yang \etal \cite{yang2019modeling} & - & 70.8 & 60.1 \\ \midrule
PCT \cite{guo2021pct} & 86.4 & 67.7 & 61.3 \\ \midrule
PVT \cite{zhang2021pvt} & 86.6 & - & 68.2 \\ \midrule
Wavelet Transformer \cite{huang2021adaptive} & 86.6 & - & - \\ \midrule
DTNet \cite{han2021dual} & 85.6 & - & - \\ \midrule
CpT \cite{kaul2021cpt} & 86.6 & 72.6 & 62.3 \\ \midrule
LFT-Net \cite{gao2022lft} & 86.2 & 76.2 & 65.2 \\ \midrule
Point-BERT   \cite{yu2021point} & 85.6 & - & - \\ \midrule
Liu \etal   \cite{liu2022group} & 86.6 & - & - \\ \midrule
Point-MAE   \cite{pang2022masked} & 86.1 & - & - \\ \midrule
PAT   \cite{cheng2021patchformer} & 86.5 & - & 68.1 \\ \midrule
3CROSSNet   \cite{han20223crossnet} & 85.3 & - & - \\ \midrule
3DMedPT \cite{yu20213d} & 84.3 & - & - \\ \midrule
MLMSPT \cite{han2021point} & 86.4 & - & 62.9\\ \midrule
Segment-Fusion \cite{thyagharajan2022segment} & - & - & 65.3 \\ \midrule
Fast Point Transformer \cite{park2021fast} & - & 77.9 & 70.3 \\ \midrule
Stratified Transformer   \cite{lai2022stratified} & 86.6 & 78.1 & 72.0\\ \bottomrule
\end{tabular}
    \label{tab:shapenet_s3dis}
    \vspacefigtext
\end{table}

\vspacesection
\subsection{Object Classification}
We show benchmark performance of transformer-based 3D object classification methods on ModelNet40 dataset \cite{wu20153d} in Table \ref{tab:modelnet}. 
This dataset is comprised of 9843 training CAD models and 2468 testing CAD models, labeled with 40 classes.
We compare the performance of the transformer-based methods to state-of-the-art non-transformer methods.
Given a CAD model, object classification methods uniformly sample a fixed number of points, and in some cases, append normal information to the input. 
In addition, some methods choose to pre-train the classification network on another dataset in a self-supervised way \cite{yu2021point,pang2022masked}. 
The quantitative evaluations on ModelNet40 show competitive results for transformer-based architectures. 
The accuracy of multiple transformer methods has exceeded that of state-of-the-art non-transformer methods.
%applications as backbone for feature extraction.
%
%MH: add citations and some justifications. 
%jean: I have commented the following sentence since it does not hold a solid idea
% Although recent extensions of non-transformer methods have shown increased performance with different training and augmentation strategies \cite{nekrasov2021mix3d,qian2022pointnext}, the performance can be further improved by the transformer architecture. 
%
%MH: this statement is a bit far fetched without sufficient justifications. It is not sufficient to relate the merits of transformers and then directly claim that they can be applied to these non-transformer methods. I remove this one and you need to add some justifications in the same sentence above. 
%Given the benefits of the transformer architecture, we believe that the newly proposed architectures have the potential to replace previous methods for better representation learning.

The classification accuracy for transformer-based methods as well as state-of-the-art non-transformer methods on the ScanObjectNN dataset \cite{scanobjectnn} is shown in Table \ref{tab:scanobjectnn}. 
This dataset represents a real-world scenario with point clouds containing background and occlusion. 
We show results with pre-training on ShapeNet dataset \cite{chang2015shapenet} and finetuning on ScanObjectNN. 
The transformer-based methods show improved performance over the non-transformer methods. 

\vspacesection
\subsection{3D Segmentation}
We show benchmark performance of various transformer-based methods for 3D segmentation in two domains: (1) object part segmentation and (2) complete scenes segmentation.
Although the task is similar in these two domains, the required data size to properly represent the shape information is usually different. 

For 3D object part segmentation, we evaluate on the widely used ShapeNet dataset \cite{chang2015shapenet}. 
It contains 14,007 training examples of point cloud objects and 2874 testing examples.
Objects are of 16 different categories, and each category is labeled with 2 to 6 parts, 
with 50 parts in total. 
For evaluation, the intersection-over-union (IoU) of each category is computed as the average IoU of all the objects in that category. 
Additionally, the instance mIoU represents the average IoU across all object instances.

Quantitative evaluation on the ShapeNet dataset is shown in Table \ref{tab:shapenet_s3dis}. %
Almost all transformer-based methods  surpass state-of-the-art point-based schemes. 
When compared to the classification task which only requires a holistic understanding of an object for category labeling, part segmentation requires local shape understanding for proper segmentation. 
This shows that the increased attention in the transformer architecture to global context supports local shape understanding.  

For scene semantic segmentation, we use the Stanford Large-Scale 3D Indoor Spaces Dataset (S3DIS) \cite{armeni2017joint}.
It is collected using Matterport scanners and contains 3D point clouds of 3 indoor areas with 271 rooms from 3 different buildings. 
For evaluation, some methods perform 6-fold validation, whereas others use Area 5 as a test set and other areas to train the model. 
Since Area 5 is different from other areas, the latter approach is more common as it better shows the method's ability to generalize to a different scene.
Quantitative comparison on the S3DIS dataset is shown in Table \ref{tab:shapenet_s3dis}. 
The quantitative evaluation shows that the early adoptions of the attention mechanism \cite{xie2018attentional} for 3D segmentation do not show significant performance gain over non-attention based methods. 
%
%Nevertheless, a few methods show improved performance, yielding better evaluation on S3DIS dataset compared to state-of-the-art non-transformer architectures. 
%
Recently, Stratified Transformer \cite{lai2022stratified} achieves significant improvement and is currently state-of-the-art on the S3DIS dataset for 3D semantic segmentation. 
The performance gain can be attributed to the efficient transformer implementation, which applies the transformer in the local context to capture fine shape information, and uses a hierarchical structure and sparse distant sampling to capture global context using a large receptive field.

\begin{table}[t]
\renewcommand{\arraystretch}{0.5}
    \centering
    \tiny
    \caption{3D object detection comparison on SUN RGB-D \cite{song2015sun} and ScanNet \cite{dai2017scannet}. We show mean AP (@IoU=0.25) for transformer-based methods and compare them to state-of-the-art non-transformer methods.}
        \begin{tabular}{p{0.15\textwidth}p{0.15\textwidth}
    p{0.1\textwidth}}
    \toprule \belowrulesepcolor{gray!30!}
\rowcolor{gray!30!} Method & SUN RGB-D mAP25 & ScanNet mAP25 \\\aboverulesepcolor{gray!30!} \midrule \belowrulesepcolor{gray!30!}
\rowcolor{gray!30!} \multicolumn{3}{c}{\textbf{Non-Transformer methods:}} \\ \aboverulesepcolor{gray!30!} \midrule
VoteNet \cite{qi2019deep} & 59.1 & 62.9 \\ \midrule
H3DNet \cite{zhang2020h3dnet} & 60.1 & 67.2 \\ \midrule \belowrulesepcolor{gray!30!}
\rowcolor{gray!30!} \multicolumn{3}{c}{\textbf{Transformer-based methods:}} \\ \aboverulesepcolor{gray!30!} \midrule
Pointformer   \cite{pan20213d} & 61.1 & 64.1 \\ \midrule
Liu \etal   \cite{liu2021group} & 63.0 & 69.1 \\ \midrule
3DETR \cite{misra2021end} & 59.1 & 65.0 \\ \midrule
Fast Point Tr.\cite{park2021fast} & - & 59.1 \\ \midrule
ARM3D \cite{lan2022arm3d} & 60.1 & 65.9 \\ \midrule
MLCVNet   \cite{xie2020mlcvnet} & 59.8 & 64.5 \\ \midrule
BrT   \cite{wang2022bridged} & 65.4 & 71.3 \\ \bottomrule
\end{tabular}
\label{tab:sunrgbd_scannet}
\vspacefigtext
\end{table}

\vspacesection
\subsection{3D Object Detection}
For indoor 3D object detection, the SUN RGB-D dataset \cite{song2015sun} provides 3D bounding box annotations to scenes from a single RGB-D frame. 
It comprises 10,335 RGB-D frames annotated with amodal and oriented bounding boxes with 37 object classes. 
Nevertheless, a standard evaluation protocol is to use only 10 common classes for training and evaluation. The training set is composed of 5285 frames whereas the testing set has 5050 frames. 

Quantitative evaluations of transformer models against state-of-the-art VoteNet \cite{qi2019deep} and H3DNet \cite{zhang2020h3dnet} methods
on the SUN RGB-D are shown in Table \ref{tab:sunrgbd_scannet}. 
%
%We evaluate transformer-based architectures against state-of-the-art VoteNet \cite{qi2019deep} and H3DNet \cite{zhang2020h3dnet} methods. 
%
While the least performing transformer model is 3DETR, 
it is a straightforward transformer encoder and decoder without exploiting recent techniques in 3D object detection. 
%
%MH: ciate other approaches
%jean: cited 3 approaches
Nevertheless, 3DETR performs as well as the VoteNet method, which has been adopted by other approaches \cite{xie2020mlcvnet,zhang2020h3dnet,chen2021s}. 
%
%For that, authors of 3DETR believe that 3DETR can serve as a building block for future research. 
%
The Bridged Transformer (BrT) \cite{wang2022bridged} achieves the best performance on the SUN RGB-D dataset. 
It benefits from the availability of RGB images and the applicability of transformers to both images and point clouds in order to bridge the learning process between the two domains. 
%
%Results show major improvement to methods such as MLCVNet which uses attention layers to supplement previous methods.

The ScanNet \cite{dai2017scannet} dataset consists of 
scenes represented by 3D meshes based on reconstruction of multiple RGB-D frames. 
The original ScanNet dataset does not include 3D bounding boxes annotations, and the hidden test set in the benchmark does not include evaluation for 3D object detection. 
Thus, the common approach is to generate axis-aligned 3D bounding boxes are enclose the provided 3D instance segmentation masks.
The training set contains 1201 scans, and the validation set of 312 scans is used for evaluation.

Performance evaluations on the ScanNet dataset are shown in Table \ref{tab:sunrgbd_scannet}. 
The Bridged Transformer (BrT) \cite{wang2022bridged} achieves the highest mAP @IoU=0.5. 
Along with the 3D scans, it uses the 25,000 frames that were used to reconstruct ScanNet scenes. 
Images are required to extract patches that are used as input to a transformer bridging the information between images and point clouds. 
For methods that do not use the additional 2D images as input, Liu \etal  \cite{liu2021group} achieve the best performance through a group-free transformer-based method. 
This method uses PointNet++ to extract local point features, and the transformer generates the bounding boxes in a global fashion.

\begin{table}[t]
\renewcommand{\arraystretch}{0.5}
    \centering
    \tiny
    \caption{Performance evaluations with state-of-the-art methods on the KITTI 3D object test set \cite{Geiger2012CVPR}. The results are reported by the mAP with 40 recall points, 0.7 IoU threshold for car, and 0.5 for others. (L: LiDAR, C: Color)}
        \begin{tabular}{p{0.076\textwidth}p{0.02\textwidth}
    p{0.1\textwidth}p{0.1\textwidth}
    p{0.1\textwidth}}
    \toprule \belowrulesepcolor{gray!30!} 
\rowcolor{gray!30!} &  & \multicolumn{1}{c}{Car (IoU=0.7)} & \multicolumn{1}{c}{Pedestrian (IoU=0.5)} & \multicolumn{1}{c}{Cyclist (IOU=0.5)} \\ \aboverulesepcolor{gray!30!} \midrule \belowrulesepcolor{gray!30!}
\rowcolor{gray!30!} Method & Input & \multicolumn{1}{c}{Easy/Mod./Hard} & \multicolumn{1}{c}{Easy/Mod./Hard} & \multicolumn{1}{c}{Easy/Mod./Hard} \\ \aboverulesepcolor{gray!30!} \midrule \belowrulesepcolor{gray!30!}
\rowcolor{gray!30!} \multicolumn{5}{c}{\textbf{Non-Transformer methods:}} \\ \aboverulesepcolor{gray!30!} \midrule
PointRCNN   \cite{shi2019pointrcnn} & L & \multicolumn{1}{c}{87.0 / 75.6 / 70.7} & \multicolumn{1}{c}{48.0 / 39.4 / 36.0} & \multicolumn{1}{c}{75.0 / 58.8 / 52.5} \\ \midrule
PV-RCNN \cite{shi2020pv} & L & \multicolumn{1}{c}{90.3 / 81.4 / 76.8} & \multicolumn{1}{c}{52.2 / 43.3 / 40.3} & \multicolumn{1}{c}{78.6 / 63.7 / 57.7} \\ \midrule
Monoflex   \cite{zhang2021objects} & C & \multicolumn{1}{c}{19.9 / 13.9 / 12.1} & \multicolumn{1}{c}{-} & \multicolumn{1}{c}{-} \\ \midrule \belowrulesepcolor{gray!30!}
\rowcolor{gray!30!} \multicolumn{5}{c}{\textbf{Transformer-based methods:}} \\ \aboverulesepcolor{gray!30!} \midrule
Pointformer   \cite{pan20213d} & L & \multicolumn{1}{c}{87.1 / 77.1 / 69.3} & \multicolumn{1}{c}{50.7 / 42.4 / 39.6} & \multicolumn{1}{c}{75.0 / 59.8 / 54.0} \\ \midrule
Voxel Trans. \cite{mao2021voxel} & L & \multicolumn{1}{c}{89.9 / 82.1 / 79.1} & \multicolumn{1}{c}{-} & \multicolumn{1}{c}{-} \\ \midrule
Sheng \etal   \cite{sheng2021improving} & L & \multicolumn{1}{c}{87.8 / 81.8 / 77.2} & \multicolumn{1}{c}{-} & \multicolumn{1}{c}{-} \\ \midrule
SA-Det3D   \cite{bhattacharyya2021sa} & L & \multicolumn{1}{c}{88.3 / 81.5 / 77.0} & \multicolumn{1}{c}{-} & \multicolumn{1}{c}{82.2 / 68.5 / 61.3} \\ \midrule
M3DETR   \cite{guan2022m3detr} & L & \multicolumn{1}{c}{90.3 / 81.7 / 77.0} & \multicolumn{1}{c}{45.7 / 39.9 / 37.7} & \multicolumn{1}{c}{83.8 / 66.7 / 59.0} \\ \midrule
Voxel Set Tr.\cite{he2022voxel} & L & \multicolumn{1}{c}{88.5 / 82.1 / 77.5} & \multicolumn{1}{c}{-} & \multicolumn{1}{c}{-} \\ \midrule
Dao \etal   \cite{dao2022attention} & L & \multicolumn{1}{c}{87.1 / 80.3 / 76.1} & \multicolumn{1}{c}{-} & \multicolumn{1}{c}{78.5 / 64.6 / 57.8} \\ \midrule
SCANet   \cite{lu2019scanet} & L+C & \multicolumn{1}{c}{76.1 / 66.3 / 58.7} & \multicolumn{1}{c}{-} & \multicolumn{1}{c}{-} \\ \midrule
MonoDETR   \cite{zhang2022monodetr} & C & \multicolumn{1}{c}{25.0 / 16.5 / 13.6} & \multicolumn{1}{c}{-} & \multicolumn{1}{c}{-} \\ \midrule
CAT-Det   \cite{zhang2022cat} & L+C & \multicolumn{1}{c}{89.9 / 81.3 / 76.7} & \multicolumn{1}{c}{54.3 / 45.4 / 41.9} & \multicolumn{1}{c}{83.7 / 68.8 / 61.5} \\ \midrule
PDV \cite{hu2022point} & L & \multicolumn{1}{c}{90.4 / 81.9 / 77.4} & \multicolumn{1}{c}{-} & \multicolumn{1}{c}{83.0 / 67.8 / 60.5} \\ \bottomrule
\end{tabular}
\label{tab:kitti}
\vspacefigtext
\end{table}

\begin{table}[t]
\renewcommand{\arraystretch}{0.5}
    \centering
    \tiny
    \caption{Performance evaluations of transformer-based methods and state-of-the-art non-transformer methods on nuScenes 3D object detection test set \cite{nuscenes}. We report the mAP for detection of 10 classes.}
        \begin{tabular}{p{0.15\textwidth}p{0.15\textwidth}p{0.1\textwidth}}
    \toprule \belowrulesepcolor{gray!30!}   
\rowcolor{gray!30!} Method    & Modality   & mAP  \\
\aboverulesepcolor{gray!30!} \midrule \belowrulesepcolor{gray!30!}
\rowcolor{gray!30!} \multicolumn{3}{c}{\textbf{Non-Transformer methods:}} \\ \aboverulesepcolor{gray!30!} \midrule
PointPillars \cite{lang2019pointpillars}                    & LiDAR          & 30.5 \\ \midrule
3DSSD   \cite{yang20203dssd}            & LiDAR          & 42.7 \\ \midrule
CBGS \cite{zhu2019class}                & LiDAR          & 52.8 \\ \midrule
FCOS3D   \cite{wang2021fcos3d}          & images         & 35.8 \\ \midrule
\belowrulesepcolor{gray!30!}
\rowcolor{gray!30!} \multicolumn{3}{c}{\textbf{Transformer-based methods:}} \\ \aboverulesepcolor{gray!30!} \midrule
Yin \etal   \cite{yin2020lidar}         & LiDAR          & 45.4 \\ \midrule
SA-Det3D   \cite{bhattacharyya2021sa}   & LiDAR          & 47.0 \\ \midrule
Pointformer   \cite{pan20213d}          & LiDAR          & 53.6 \\ \midrule
Dao \etal   \cite{dao2022attention}     & LiDAR          & 47.0 \\ \midrule
VISTA   \cite{deng2022vista}            & LiDAR          & 63.0 \\ \midrule
Transfusion   \cite{bai2022transfusion} & LiDAR + images & 68.9 \\ \midrule
PETR \cite{liu2022petr}                 & images         & 43.4 \\ \midrule
DETR3D   \cite{wang2022detr3d}          & images         & 41.2 \\ \midrule
Yuan \etal   \cite{yuan2021temporal}    & images         & 50.5 \\ \bottomrule
\end{tabular}
\label{tab:nuscenes}
\vspacefigtext
\end{table}

\begin{table}[t]
\renewcommand{\arraystretch}{0.5}
\centering
\tiny
\caption{Performance evaluations of point cloud completion  on Completion3D \cite{Tchapmi2019com3d} and PCN \cite{Yuan2018pcn}. L2 Chamfer distance is used as metric on  Completion3D, while L1 Chamfer distance is used as metric on PCN. }
    \begin{tabular}{p{0.15\textwidth}p{0.12\textwidth}p{0.12\textwidth}}
\toprule \belowrulesepcolor{gray!30!}
\rowcolor{gray!30!} Method & Completion3D: Avg. ↓ & PCN: Avg. ↓\\ \aboverulesepcolor{gray!30!} \midrule \belowrulesepcolor{gray!30!}
\rowcolor{gray!30!} \multicolumn{3}{c}{\textbf{Non-Transformer methods:}} \\ \aboverulesepcolor{gray!30!} \midrule
PCN \cite{Yuan2018pcn} &  18.22 & 9.64 \\ \midrule
PMP-Net \cite{Wen2021PMPNet} &   9.23 & 8.66\\ \midrule \belowrulesepcolor{gray!30!}
\rowcolor{gray!30!} \multicolumn{3}{c}{\textbf{Transformer-based methods:}} \\ \aboverulesepcolor{gray!30!} \midrule
PoinTr \cite{Yu2021PoinTr} & - & 8.38\\ \midrule
Snowflakenet \cite{Xiang2021SnowflakeNet} &  7.60  & 7.21\\ \midrule
PointAttN \cite{Wang2022PointAttN} & 6.63 & 6.86\\ \midrule
Wang \etal \cite{Wang2022localdisp} & 6.64 & 7.96 \\ \midrule
% PCTMA-Net \cite{Lin2021PCTMANet} & 9.48 & - \\ \midrule
\end{tabular}
\label{exp:completion3d}
\vspacefigtext
\end{table}

For outdoor 3D object detection, the KITTI dataset \cite{Geiger2012CVPR} is one of the most widely used datasets for autonomous driving and provides 3D object detection annotations. 
It contains 7481 training samples and 7518 testing samples and uses standard average precision (AP) on easy, moderate, and hard difficulties. 
The performance evaluations on KITTI benchmark are shown in Table \ref{tab:kitti}. 
We also show the input used in each method, which can either be LiDAR only, RGB only, or both. 
Existing methods that rely on RGB alone under-perform the ones that make use of the LiDAR information. 
Since the task is to localize objects in the 3D scene, RGB alone lacks necessary 3D information required for accurately placing the bounding boxes. 
Nevertheless, the transformer-based architecture MonoDETR \cite{zhang2022monodetr} outperforms Monoflex \cite{zhang2021objects}. 
For LiDAR input, PDV \cite{hu2022point} achieves the best performance  on the ``easy'' car category. It makes use of a self-attention module to capture long-range dependencies of grid points, where features are computed using 3D sparse convolutions. 
For the moderate and hard difficulties, the Voxel Transformer \cite{mao2021voxel} and the Voxel Set Transformer \cite{he2022voxel} achieve the best performance, benefiting from the integration of the attention mechanism into the feature extraction module.

Another widely used outdoor dataset for 3D object detection is nuScenes \cite{nuscenes}, which contains 1000 scenes that include heavy traffic and challenging driving situations. 
We show detection results for various methods in Table \ref{tab:nuscenes}. 
We compare transformer models using depth from LiDAR and/or images to state-of-the-art non-transformer methods. 
The transformer methods show improved performance for both the LiDAR-based and image-based 3D object detection. 
The best performance is achieved with the Transfusion method \cite{bai2022transfusion} in which a transformer fuses information from both the depth and color modalities. 
The results show that the transformer is capable of learning the complementary information from both modalities.

\vspacesection
\subsection{3D Point Cloud Completion}
As discussed earlier, Completion3D \cite{Tchapmi2019com3d} and PCN \cite{Yuan2018pcn} are usually used for 3D point completion. 
Completion3D dataset has 30,958 point cloud models in 8 categories. 
The partial and complete point clouds have the same size of 2,048 $\times$ 3. 
The partial 3D point clouds are back-projection of depth images from 3D space, and  L2 Chamfer distance is used for performance evaluation on Completion3D \cite{Tchapmi2019com3d}. 
Table \ref{exp:completion3d} (middle column) shows the point completion performance of some non-transformer and transformer-based methods on Completion3D dataset \cite{Tchapmi2019com3d}. 
PCN \cite{Yuan2018pcn} and PMP-Net \cite{Wen2021PMPNet} are non-transformer methods whereas PoinTr \cite{Yu2021PoinTr}, Snowflakenet \cite{Xiang2021SnowflakeNet}, PointAttN \cite{Wang2022PointAttN}, and Wang \etal \cite{Wang2022localdisp} are transformer-based methods. 
%
%Compared to non-transformer methods, the transformer-based methods usually have better performance. 
Clearly, methods based on transformers perform better for this task using Completion3D. 
Among these transformer-based methods, PointAttN \cite{Wang2022PointAttN} and Wang \etal \cite{Wang2022localdisp} have the smallest L2 Chamfer distance (\textit{i.e.,} 6.63 and 6.64), where they employ a transformer to extract global and local information for improved performance.

PCN \cite{Yuan2018pcn} dataset has 30,974 point cloud models in 8 categories. 
The partial point clouds have less than 20,48 points, while the complete point clouds have 16,384 points.
In the following experiments, L1 Chamfer distance is used for performance evaluation on PCN dataset \cite{Yuan2018pcn}. 
Table \ref{exp:completion3d} (right column) shows the point completion performance on PCN \cite{Tchapmi2019com3d}. 
%
% PCN \cite{Yuan2018pcn} and PMP-Net \cite{Wen2021PMPNet} are non-transformer methods, where as PoinTr \cite{Yu2021PoinTr}, Snowflakenet \cite{Xiang2021SnowflakeNet}, PointAttN \cite{Wang2022PointAttN}, and Wang \etal \cite{Wang2022localdisp} are transformer-based methods. 
%
%Compared to non-transformer methods, the transformer-based methods usually have better performance. 
%
Among these transformer-based methods, Snowflakenet \cite{Xiang2021SnowflakeNet} and  PointAttN \cite{Wang2022PointAttN}  have the smallest L1 Chamfer distance (\textit{i.e.,} 7.21 and 6.86).

\begin{table}[t]
\renewcommand{\arraystretch}{0.5}
    \centering
    \tiny
    \caption{3D pose estimation performance comparison  on
Human3.6M \cite{Ionescu2014Human36m} under Protocols 1\&2 where 2D pose detection is used as input.}
        \begin{tabular}{p{0.15\textwidth}p{0.12\textwidth}p{0.12\textwidth}}
    \toprule \belowrulesepcolor{gray!30!}
\rowcolor{gray!30!} Method  & Protocol 1: Avg. ↓  & Protocol 2: Avg. ↓  \\ \aboverulesepcolor{gray!30!}  \midrule \belowrulesepcolor{gray!30!}
\rowcolor{gray!30!} \multicolumn{3}{c}{\textbf{Non-Transformer methods:}} \\ \aboverulesepcolor{gray!30!} \midrule
UGCN \cite{Wang2020UCGN} &   45.6 & 35.5\\ \midrule
Chen \textit{et al.} \cite{Chen2021lAA3D} &   44.6 & 35.6 \\
\midrule \belowrulesepcolor{gray!30!}
\rowcolor{gray!30!} \multicolumn{3}{c}{\textbf{Transformer-based methods:}} \\ \aboverulesepcolor{gray!30!} \midrule
METRO \cite{Lin2021METRO} &   54.0 &  36.7\\ \midrule
PoseFormer \cite{Zheng2021PoseFormer} & 44.3 & 34.6 \\ \midrule
% GraFormer \cite{Zhao2022GraFormer} & 58.7 & -\\ \midrule
CrossFormer \cite{Hassanin2022CrossFormer} & 43.7 & 34.3 \\ \midrule
STE \cite{Li2022STE} & 43.7 & 35.2\\ \midrule
P-STMO \cite{Shan2022P-STMO} & 44.1 & 34.4\\ \midrule
MHFormer \cite{Li2022MHFormer} & 43.0 & -\\ \midrule
MixSTE \cite{Zhang2022MixSTE} & 42.4  & 33.9\\ \bottomrule
% \midrule \belowrulesepcolor{gray!30!}
% \rowcolor{gray!30!} Protocol 2  & Avg. ↓ \\ \aboverulesepcolor{gray!30!} \midrule \belowrulesepcolor{gray!30!}
% \rowcolor{gray!30!} \multicolumn{2}{c}{\textbf{Non-Transformer methods:}} \\ \aboverulesepcolor{gray!30!} \midrule
% UGCN \cite{Wang2020UCGN} &   35.5 \\  \midrule
% Chen \textit{et al.} \cite{Chen2021lAA3D} &   35.6 \\ \midrule \belowrulesepcolor{gray!30!}
% \rowcolor{gray!30!} \multicolumn{2}{c}{\textbf{Transformer-based methods:}} \\ \aboverulesepcolor{gray!30!} \midrule
% PoseFormer \cite{Zheng2021PoseFormer} & 34.6 \\ \midrule
% CrossFormer \cite{Hassanin2022CrossFormer} & 34.3 \\ \midrule
% STE \cite{Li2022STE} & 35.2 \\  \midrule
% P-STMO \cite{Shan2022P-STMO} & 34.4 \\ \midrule
% MixSTE \cite{Zhang2022MixSTE} & 33.9 \\ \bottomrule
\end{tabular}
\label{exp:human}
\vspacefigtext
\end{table}

\begin{table}[t]
\renewcommand{\arraystretch}{0.5}
    \centering
    \tiny
    \caption{3D pose estimation performance comparison on MPI-INF-3DHP \cite{Mehta20173DHP}.}
        \begin{tabular}{p{0.15\textwidth}p{0.08\textwidth}
    p{0.08\textwidth}p{0.08\textwidth}}
    \toprule \belowrulesepcolor{gray!30!}
\rowcolor{gray!30!} Method  & PCK ↑ & AUC ↑ & MPJPE ↓ \\ \aboverulesepcolor{gray!30!} \midrule \belowrulesepcolor{gray!30!}
\rowcolor{gray!30!} \multicolumn{4}{c}{\textbf{Non-Transformer methods:}} \\ \aboverulesepcolor{gray!30!} \midrule
Chen \textit{et al.} \cite{Chen2021lAA3D} &   87.9 & 54.0 & 78.8 \\
\midrule \belowrulesepcolor{gray!30!}
\rowcolor{gray!30!} \multicolumn{4}{c}{\textbf{Transformer-based methods:}} \\ \aboverulesepcolor{gray!30!} \midrule
PoseFormer \cite{Zheng2021PoseFormer} & 88.6 &  56.4 & 77.1 \\ \midrule
CrossFormer \cite{Hassanin2022CrossFormer} &  89.1 & 57.5 & 76.3\\ \midrule
P-STMO \cite{Shan2022P-STMO} &  97.9& 75.8& 32.2\\ \midrule
MHFormer \cite{Li2022MHFormer} &  93.8 & 63.3 & 58.0
\\ \midrule
MixSTE \cite{Zhang2022MixSTE} &  94.2 & 63.8 & 57.9
\\ \bottomrule
\end{tabular}
\label{exp:3DHP}
\vspacefigtext
\end{table}

%MH: same issue with Section 5.5
% add discussion of best methods
\vspacesection
\subsection{3D Pose estimation}
Human3.6M \cite{Ionescu2014Human36m} and MPI-INF-3DHP \cite{Mehta20173DHP} are two widely used 3D pose estimation datasets for 3D pose estimation. 
The Human3.6M dataset \cite{Ionescu2014Human36m} contains 3.6 million video frames recorded in 4 different views of the indoor scene. 
There are 17 action categories performed by 11 different actors. 
Two metrics (MPJPE and P-MPJPE) are used for performance evaluation. 
The mean per joint position error (MPJPE) metric,  denoted as Protocol 1, is calculated as the averaged Euclidean distance  in millimeters between the predicted and ground-truth joints.
The P-MPJPE metric, denoted as Protocol 2, is the MPJPE after the post-processing of rigid alignment. 
Table \ref{exp:human} shows the 3D pose estimation performance of some non-transformer and transformer-based methods on Completion3D \cite{Tchapmi2019com3d}. 
Compared to non-transformer schemes, most transformer-based methods usually have better performance under both Protocol 1 and Protocol 2. 
Among these transformer-based methods, MixSTE \cite{Zhang2022MixSTE} has the best performance with an average MPJPE of 42.4 and 33.9 under Protocol 1 and Protocol 2. MixSTE consists of a temporal transformer and a spatial transformer, where the two transformers are performed in an  alternating style. The temporal transformer is to learn the temporal relation of each joint, while the spatial transformer is to build the spatial correlation of different joints. 

The MPI-INF-3DHP dataset \cite{Mehta20173DHP} consists of 1.3 million frames recorded in both  indoor and outdoor scenes.
There are 8 action categories performed by 8 different actors. Table \ref{exp:3DHP} shows the results of some methods. Three metrics (MPJPE, PCK, and AUC) are used for performance evaluation. 
The PCK metric is a 3D extension of the percentage of correct keypoints with a threshold of 150mm, while AUC is calculated with a range of PCK thresholds. 
Among these transformer-based methods, P-STMO \cite{Shan2022P-STMO} has the best performance, with PCK of 97.9, AUC of 75.8, and MPJPE of 32.2. P-STMO employs the self-supervised pre-training transformer for pose estimation.

\vspacesection
\section{Discussion and Conclusion}
Integrating the transformer into the pipeline of 3D applications has been shown to be effective in numerous areas.
Given the competitive performance on multiple datasets, the transformer was demonstrated to be an adequate replacement for convolution and multi-layer perceptron operations, thanks to its ability to learn long-range dependencies.
Nonetheless, a generic transformer backbone for 3D processing is still missing. 
Unlike image processing with transformers which has focal methods that many other methods rely on \cite{dosovitskiy2020image,liu2021swin}, most transformer-based 3D methods use different transformer designs and integration. 
It is of great interest to develop a general-purpose transformer method that processes point clouds and learns rich features on the local and global scale. 
The transformer is required to learn fine shape information, and at the same time operate on a scene global scale to make use of the scene context. 

Additionally, most transformer-based 3D methods sub-sample 3D data to a fixed size input. 
A fixed input size is the norm in images considering the given number of pixels, but 3D input size usually varies. 
Therefore, a proper sampling strategy that (1) preserves the potential to learn fine information, (2) maintains similar object structure irrespective of the scene size, and (3) generates a feasible data size for transformer processing, is essential. 
Although NLP processes inputs of varying sizes, words in sentences are all tokenized, and input information is not lost. On the other hand, 3D input sampling leaves out potentially important information and is dependent on the scene size.
A sampling strategy worth exploring is to use a data-driven approach. 
For example, a small sub-sample of the input can be used as seeds for a flexible and data-driven sampling. 
This is similar to previous input aggregation methods, but sampling would be adapted to input information, and original information would always be accessible. 

The choice of the position embedding is also essential. 
The position information in the 3D field not only orders the information sequences but is also the main feature for understanding shape information.
While most 3D vision methods use the 3D position as part of the input to the transformer, this information might require prior encoding and might not be explicitly available after multiple feature extraction layers. 
Therefore, a proper position embedding would benefit the learning task. The choice of the position embedding should help models learn translation invariance and retain context information from other scene elements.  

%MH: check this paragraph 
Existing transformer models rely on data augmentation in the training process.  
Although numerous transformer methods use off-the-shelf 3D augmentation methods, several 3D augmentation techniques have shown significant improvement over common non-transformer methods \cite{nekrasov2021mix3d,qian2022pointnext}. % cited 3d mix and pointnext
A similar trend could also benefit transformer based methods, with the utilization of 3D-specific augmentations that are well suited for training transformers.

Compared to NLP or 2D image processing, 3D vision rely less on pre-training. 
This is also observed in transformer-based methods, where pre-training is seldom used. Large-scale fully-supervised or self-supervised pre-training has helped improve performance as well as the robustness of 2D models. 
Similar pre-training models would also benefit the 3D domain. 
This can either be done with learning on a fully labeled large-scale dataset, akin to ImageNet in 2D, or with self-supervised learning.

% if have a single appendix:
%\appendix[Proof of the Zonklar Equations]
% or
%\appendix  % for no appendix heading
% do not use \section anymore after \appendix, only \section*
% is possibly needed

% use appendices with more than one appendix
% then use \section to start each appendix
% you must declare a \section before using any
% \subsection or using \label (\appendices by itself
% starts a section numbered zero.)
%

% \appendices
% \section{}
% Appendix one text goes here.

% you can choose not to have a title for an appendix
% if you want by leaving the argument blank
% \section{}
% Appendix two text goes here.

% % use section* for acknowledgment
% \ifCLASSOPTIONcompsoc
%   % The Computer Society usually uses the plural form
%   \section*{Acknowledgments}
% \else
%   % regular IEEE prefers the singular form
%   \section*{Acknowledgment}
% \fi

% The authors would like to thank...

% Can use something like this to put references on a page
% by themselves when using endfloat and the captionsoff option.
\ifCLASSOPTIONcaptionsoff
  \newpage
\fi

% trigger a \newpage just before the given reference
% number - used to balance the columns on the last page
% adjust value as needed - may need to be readjusted if
% the document is modified later
%\IEEEtriggeratref{8}
% The "triggered" command can be changed if desired:
%\IEEEtriggercmd{\enlargethispage{-5in}}

% references section

% can use a bibliography generated by BibTeX as a .bbl file
% BibTeX documentation can be easily obtained at:
% http://mirror.ctan.org/biblio/bibtex/contrib/doc/
% The IEEEtran BibTeX style support page is at:
% http://www.michaelshell.org/tex/ieeetran/bibtex/

\vspacesection
{
\bibliographystyle{IEEEtran}
% argument is your BibTeX string definitions and bibliography database(s)
%\bibliography{IEEEabrv,../bib/paper}
\bibliography{references}
}

%
% <OR> manually copy in the resultant .bbl file
% set second argument of \begin to the number of references
% (used to reserve space for the reference number labels box)
% \begin{thebibliography}{1}

% \end{thebibliography}

% biography section
% 
% If you have an EPS/PDF photo (graphicx package needed) extra braces are
% needed around the contents of the optional argument to biography to prevent
% the LaTeX parser from getting confused when it sees the complicated
% \includegraphics command within an optional argument. (You could create
% your own custom macro containing the \includegraphics command to make things
% simpler here.)
%\begin{IEEEbiography}[{\includegraphics[width=1in,height=1.25in,clip,keepaspectratio]{mshell}}]{Michael Shell}
% or if you just want to reserve a space for a photo:

% \begin{IEEEbiography}{Michael Shell}
% Biography text here.
% \end{IEEEbiography}

% % if you will not have a photo at all:
% \begin{IEEEbiographynophoto}{John Doe}
% Biography text here.
% \end{IEEEbiographynophoto}

% % insert where needed to balance the two columns on the last page with
% % biographies
% %\newpage

% \begin{IEEEbiographynophoto}{Jane Doe}
% Biography text here.
% \end{IEEEbiographynophoto}

% You can push biographies down or up by placing
% a \vfill before or after them. The appropriate
% use of \vfill depends on what kind of text is
% on the last page and whether or not the columns
% are being equalized.

%\vfill

% Can be used to pull up biographies so that the bottom of the last one
% is flush with the other column.
%\enlargethispage{-5in}

% that's all folks
\end{document}